\documentclass[10pt,twocolumn,letterpaper]{article}

\usepackage{iccv}
\usepackage{times}
\usepackage{epsfig}
\usepackage{graphicx}
\usepackage{amsmath}
\usepackage{amssymb}
\usepackage{multirow}
\usepackage{subcaption}
\usepackage[dvipsnames]{xcolor}
\usepackage{tabularx}
\usepackage{booktabs}
\usepackage[font=small]{caption}
\usepackage{slashbox}
\usepackage[pagebackref=true,breaklinks=true,colorlinks,bookmarks=false]{hyperref}

 \iccvfinalcopy % *** Uncomment this line for the final submission

\ificcvfinal\fi

\begin{document}

%%%%%%%%% TITLE
\title{A Light Stage on Every Desk}

\author{
Soumyadip Sengupta\hspace{7mm}
Brian Curless\hspace{7mm}
Ira Kemelmacher-Shlizerman\hspace{7mm}
 Steve Seitz\\
University of Washington \\
}

\maketitle
\begin{abstract}
\vspace{-0.5em}
Every time you sit in front of a TV or monitor, your face is actively illuminated by time-varying patterns of light.  This paper proposes to use this time-varying illumination for synthetic relighting of your face with any new illumination condition. In doing so, we take inspiration from the {\em light stage} work of Debevec et al. \cite{debevec2000acquiring}, who first demonstrated the ability to relight people captured in a controlled lighting environment.  Whereas existing light stages require expensive, room-scale spherical capture gantries and exist in only a few labs in the world, we demonstrate how to acquire useful data from a normal TV or desktop monitor. Instead of subjecting the user to uncomfortable rapidly flashing light patterns, we operate on images of the user watching a YouTube video or other standard content. 
We train a deep network on images plus monitor patterns of a given user and learn to predict images of that user under any target illumination (monitor pattern). Experimental evaluation shows that our method produces realistic relighting results. Video results are available at {\scriptsize \url{grail.cs.washington.edu/projects/Light_Stage_on_Every_Desk/}}.

\end{abstract}
%\vspace{-1em}
\vspace{-1.0em}
\section{Introduction}
\label{sec:intro}
\vspace{-0.5em}

A light stage, first introduced by Debevec \etal \cite{debevec2000acquiring}, is an instrument to capture and render human subjects under almost any illumination condition.  Most light stages consist of room-scale, spherical arrays of brightly-flashing colored lights and cameras.  They are used widely for movie special effects \cite{debevec2002lighting,alexander2010digital,alexander2013digital,wenger2005performance,peers2007post}, volumetric media \cite{zhang2021neural,guo2019relightables}, presidential portraits \cite{metallo2015scanning}, and to provide rich data for training computer vision relighting algorithms \cite{sun2019single,zhang2020portrait,sun2020light,legendre2020learning,nestmeyer2020learning,wang2020single,mallikarjun2020monocular}. 

Unfortunately, very few researchers have access to a light stage, as only a few exist in the world.  Furthermore, light stage datasets (necessary to train relighting algorithms) are not publicly available.  The goal of this paper is to democratize light stage capture and the development of new relighting algorithms.

Our key insight is that the simple act of watching video on a monitor or TV sends patterns of light across your face.  By analyzing these patterns, we can achieve face relighting in a manner similar to a light stage.  
The ability to capture “light stage” data from displays we use everyday and from the  content we already watch dramatically broadens the access and applicability of light stage and relighting research.  For example, imagine improving how your face is lit in a video call, by analyzing the changes in your facial appearance via monitor lighting over the last 10 minutes.  However, it comes with some new challenges, namely:\\
\noindent1. \textit{passive vs. active lighting}:  we wish to avoid forcing the user to watch flashing light patterns, and instead operate on the natural, time-varying content people normally watch.\\
\noindent2. \textit{motion}:  user head motion combined with time-varying light patterns complicate the registration problem (in contrast to traditional light stage techniques, which send white frames to enable optical flow).\\
\noindent3. \textit{field of view}:  the monitor provides only frontal lighting from a limited field of view.\\
\noindent4. \textit{brightness}:  we are limited by monitor brightness relative to room lighting.\\
\noindent5. \textit{near-field}:  the monitor is a proximal source, and does not model distant lighting.

In this paper, we address the first two of these challenges, which effectively transform every desktop monitor into a light stage.  We show that the resulting data enables new relighting algorithms and applications such as improving face lighting for video calls.  While we believe future research can improve the operating range, our approach works best for dimly lit rooms or bright monitors.

To this end, we propose a deep network that takes as input a face image and corresponding source pattern (image on monitor), and produces an image of the same face under a desired target pattern.
This network is trained using imagery of a moving person watching a (known) monitor video.  To normalize for user motion, we find pairs of images where the face is approximately in the same pose, but with different monitor patterns.  We then train the network to relight the first image of the pair with the pattern of the other, using perceptual, cycle consistency and adversarial losses based on PatchGAN \cite{zhu2017unpaired}.
We handle lighting as a style, and use Adaptive Instance Normalization (AdaIN) \cite{huang2017arbitrary} to infuse lighting information into our deep network. The parameters of AdaIN are learnt from input and target lighting pattern with a multi-layer perceptron.  We show state-of-the-art results on a variety of scenes and captures. 

In the following section we define the problem and reference related work, as well as fold related work into the subsequent sections. 
%\vspace{-0.5em}
\section{Problem Definition}
%\vspace{-0.5em}
\label{sec:problem}
The {\em performance relighting} problem \cite{wenger2005performance} is to capture video of a moving subject that can be synthetically re-lit with new target lighting.
Specifically, given a set of input images of the subject 
$\mathbf{I} = [I^i...I^N]$, input light maps
$\mathbf{L} = [L^1...L^N]$, and target light maps 
$\mathbf{L}_t = [L^{1}_{t}...L^{N}_{t}]$,
predict how each input image $I^i$ appears under target $L^{i}_{t}$, to produce re-lit image sequence 
$\mathbf{I}_t = [I^{1}_{t}...I^{N}_{t}]$.
(The representation of $L^i$ is implementation-dependent.)
While performance relighting is limited to re-illuminating the specific person who was captured, many solutions enable very detailed, view-dependent effects, and movie-quality output. 

In contrast, {\em portrait relighting} methods operate on {\em a single image of a subject} $I$, transforming it to a re-lit image $I_t$ under a target light $L_t$.  
While earlier deep net methods leveraged synthetic data
\cite{sengupta2018sfsnet,zhou2019deep}, most state-of-the-art techniques now rely on light stage data for portrait relighting \cite{sun2019single,zhang2020portrait,sun2020light,legendre2020learning,nestmeyer2020learning,wang2020single,mallikarjun2020monocular}. 
Portrait relighting methods require training data spanning a wide range of people.

In this paper, we focus on the performance relighting problem where the illumination comes from a conventional monitor or TV.  Prior research has also considered using monitors~\cite{Chuang:2000:EME,Zongker:1999:EMC} or projection screens~\cite{schechner2007multiplexing} as light sources to enable relighting of static objects or for 3D reconstruction with Photometric Stereo \cite{schindler2008photometric}. In contrast, our approach is suitable for relighting videos, for example to improve 
facial appearance in YouTube or video calls.  
A second application is to normalize facial lighting for game players or other users in front of monitors.  As inverse lighting methods progress, it may become possible to decode screen content from face video \cite{nishino2004eyes};  performance relighting provides a way to foil such privacy attacks.

\begin{figure*}[t]
	\centering
	\includegraphics[width=.98\textwidth]{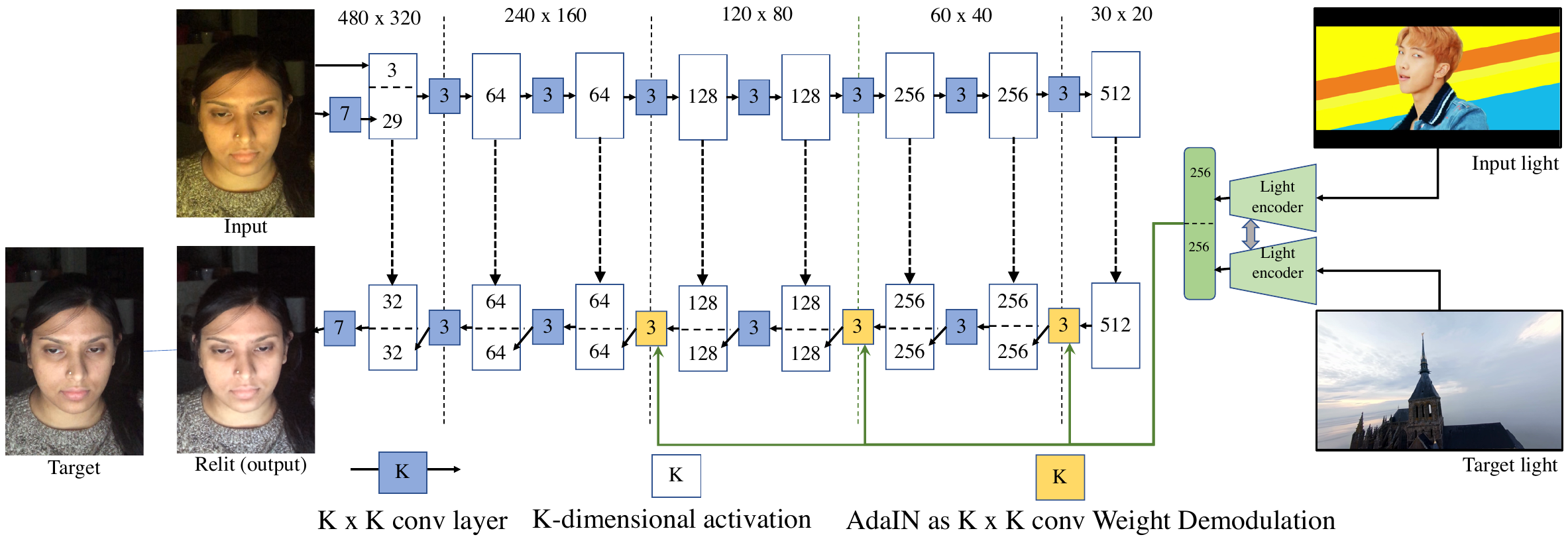}    \vspace{-0.5em}
	\caption{\small Our proposed architecture consists of an U-Net and a light encoder which estimates a $256$ dimensional latent subspace for each of the input and target lights (monitor images). It is then used to demodulate the weights of 3x3 convolutional kernels, similar to Adaptive Instance Normalization (AdaIN) layer, as proposed in StyleGANv2 \cite{Karras2019stylegan2}. The network is trained with L1, perceptual, adversarial, and cycle consistency loss functions.}
	\vspace{-1.5em}
	\label{fig:net}
\end{figure*}

%\vspace{-0.5em}
\section{Monitor as a Light Stage}
%\vspace{-0.5em}

In our setup, the subject sits in front of a monitor while being passively (and primarily) illuminated by screen content, e.g., a YouTube video.    Figure 1(b) illustrates this simple scenario.  We record both the frames shown on the screen and a synchronized webcam view of the subject.  The result is a set of screen (lighting) images $\mathbf{L}$ 
and corresponding images of the subject $\mathbf{I}$.

Later, while still capturing the screen and subject, we seek to relight the subject. This means at test time, we have source lighting $L_s$, source image $I_s$, and target lighting $L_t$, and seek to produce relit image $\hat{I}_t$.

\subsection{Linear baseline}
 \label{sec:ls}

At first glance the problem and solution seem straightforward.  Indeed, if the subject is perfectly still and we follow the light stage approach of illuminating one pixel (or non-overlapping group of pixels) at a time, then we can simply solve for the weighted combination of input lights that produce the target lighting:
\begin{equation}
\vspace{-0.5em}
  L_t =  \sum_i w_i L^i = \mathbf{L} w,
  \vspace{-0.5em}
\end{equation}
where $w$ is the weight vector $[w_1...w_N]^T$.

Given the linearity of incoherent light transport (and after linearizing for camera response), we can apply those same weights to the input images to get the relit image:
\begin{equation}
\vspace{-0.5em}
  \hat{I}_t = \sum_i w_i I^i = \mathbf{I} w.
 \vspace{-0.5em}
\end{equation}

Solving for $w$ is trivial if pixel groups are turned on one at time ($\mathbf{L}$ is identity matrix), but for aribtrary lighting sequences, we can compute the least-squares optimal weights:
\begin{equation}
\vspace{-0.5em}
w = \mathbf{L}^+ L_t 
\end{equation}
where $\mathbf{L}^+$ is the pseudoinverse of $\mathbf{L}$.  Notice how, in this linear formulation, we don’t need the source lighting $L_s$, or even the source image $I_s$, to generate the result $\hat{I}_t$.

The critical assumption in this linear method is that the subject is perfectly still, which is not true in practice.  Light stage approaches typically overcome this limitation by asking the subject to hold (fairly) still, or by using a very high frame rate camera, and then introducing white frames (all lights on) during capture. This allows optical flow to register all imagery to compensate for small subject motion.  In our setting, we work with uncontrolled imagery (no white frames injected) and subjects who move their heads in natural ways (not required to hold still) while watching that imagery in front of a conventional camera.

\subsection{Our approach}
\vspace{-0.5em}

To address the challenge of subject's natural head motion while utilizing the lighting and image pairs at test time, we train a deep network $G$ to produce the desired, relit image:
\begin{equation}
\vspace{-0.5em}
  \hat{I}_t = G(I_s,L_s,L_t)
\end{equation}

We take inspiration from single-image portrait relighting techniques, which do not depend on $L_s$, and generalize to include $L_s$ as input (Section~\ref{sec:architecture}) and handle head motion in the data (Section~\ref{sec:training}).  The motion is especially challenging because we have a continuous stream of uncontrolled lighting and capture, but no ground-truth supervision in the form of pixel-aligned source and target image pairs, i.e., $I_s$ and $I_t$.  We train instead, on all pairs of images with {\em roughly similar} poses and design network losses that do not require exact alignment to produce good results.  Note that single-image methods \cite{sun2019single,wang2020single,nestmeyer2020learning} are trained on carefully registered data to generalize and relight any image with unknown source lighting. Our approach fits to a single capture of an individual without registered data, but with source lighting available at test time.
\vspace{-0.5em}
\subsubsection{Network architecture}
\label{sec:architecture}
\vspace{-0.5em}

We employ a U-net architecture (shown in Figure \ref{fig:net}), similar to one used by Sun \textit{et al.}~\cite{sun2019single}.
The U-Net encoder downsamples the image four times, each time by a factor of two, to obtain the latent feature space. The U-Net decoder expands these features with bi-linear upsampling followed by 3$\times$3 convolution. Skip connections from the encoder are also used at each step of the decoder. Activation block involves pixel normalization followed by learnable ReLUs.

Recent state-of-the-art relighting methods have incorporated lighting in different ways.  Sun \textit{et al.}~\cite{sun2019single} simply concatenate the environment map as channels in the latent subspace, while Wang~\textit{et al.}~\cite{wang2020single} modulate the image features at each level of the U-Net decoder producing better results.  Our approach is inspired by the latter, though our overall architecture is quite different as Wang~\textit{et al.} need normal and albedo maps for training.

As noted earlier, unlike prior single-image relighting methods, we have as input the source lighting $L_s$, in addition to the target lighting $L_t$. We employ a single `Light Encoder' (a multi-layer perceptron (MLP)) that maps a lighting image to a low-dimensional code ($d=256$).  We apply the Light Encoder to $L_s$ and to $L_t$, concatenate the two codes. These codes are then individually used to demodulate the weights of convolutional kernels at all decoder levels, except the highest resolution. MLPs are used for predicting the weights and biases needed for demodulating the convolutional kernels. We empirically observe that this implementation produces less artifacts than using native Adaptive Instance Normalization (AdaIn)~\cite{huang2017arbitrary} layers.

\vspace{-0.5em}
\subsubsection{Training}
\label{sec:training}
\vspace{-0.5em}

Due to head motion in the dataset, we do not have pixel-aligned images of the subject under different lighting conditions for supervised network training.  Instead, we find pairs of images with similar (though not identical) poses.  In particular, we apply a face parsing network pre-trained on FFHQ~\cite{yu2018bisenet} to every image and then find every pair of images with intersection-over-union between the face segments to be greater than 92\%.  Each pair $(I_s, I_t)$ along with corresponding lighting $(L_s,L_t)$ is used to supervise training of the network $G$.  

We define a set of losses to encourage the generated image $\hat{I}_t=G(I_s,L_s,L_t)$ to match supervision $I_t$.  First, we define L1 and perceptual losses:
\begin{align}
\vspace{-0.5em}
    \mathcal{L}_{L1} &= \|I_t - \hat{I}_t\|_1, \label{eq:l2}\\
    \mathcal{L}_{P} &= \|F(I_t) - F(\hat{I}_t)\|_2. \label{eq:p}
\vspace{-0.5em}
\end{align}
where, $F(\cdot)$ is a multi-scale, deep feature extractor~\cite{zhang2018unreasonable}.  

For regularization, employ a cycle consistency loss:
\begin{align}
\vspace{-0.5em}
    \mathcal{L}_{C} &= \|I_s - G(\hat{I}_t,L_t,L_s)\|_1. \label{eq:reg}
\vspace{-0.5em}
\end{align}

The aforementioned losses often result in blurry re-lit images due to misalignment. Registering the pairs with optical flow is difficult given the lighting differences between them. Similarly, facial keypoint detection is inaccurate in the relatively low-light regime of monitor illumination, making it difficult to obtain accurate warping. To address this blur, we incorporate an adversarial loss in which  patches of the relit image $\hat{I}_t$ are compared against patches in $I_t$. We use the PatchGAN discriminator~\cite{zhu2017unpaired} $D$ which classifies every 70$\times$70 patches as real or fake. We use the LS-GAN~\cite{mao2017least} framework to train our generator $G$ and discriminator $D$. For the generator update, we minimize over the parameters of $G$:

\begin{equation}
\vspace{-0.5em}
\begin{split}
\min_{G} ~ \lambda_{L1} \mathcal{L}_{L1} + \lambda_{P} \mathcal{L}_{P} + \lambda_{C} \mathcal{L}_{C} + \lambda_{D} (D(\hat{I}_t) -1)^2,
\end{split}
\vspace{-0.5em}
\end{equation}
where each loss depends on $\hat{I}_t=G(\cdot)$.  For the discriminator update, we minimize over the parameters of $D$:
\begin{equation}
\vspace{-0.5em}
    \min_{D} (D(I_t) - 1)^2 + (D(\hat{I}_t))^2.
\end{equation}

\textbf{Details.} Each input image resolution is 480$\times$320, and the lighting resolution is 18$\times$32. The input images are cropped to the subject's head to limit the effect of the background. We use $\lambda_{L1}=1$, $\lambda_{P}=0.1$, $\lambda_{C}=0.5$ and $\lambda_{D}=0.1$. We train the generator and  discriminator with the Adam optimizer, with learning rate of $10^{-3}$ and $10^{-6}$, and a batch-size of 1.

% \vspace{-0.5em}

\section{Experimental Evaluation}
\label{sec:exp-sec}

\textbf{Capture} Our setup consists of a  computer monitor and a camera in a dimly lit room (illustrated in Fig 1b). A person watches a YouTube video on the monitor while their appearance is captured by the camera.  In  our experiments we either turned off the light in the room, or kept low light, and assumed the user is relatively close to the monitor (about $30cm$) to obtain good signal to noise ratio. We tested with both 32" (LG) and 27" (Acer) monitor sizes.  The larger monitor is preferred to allow more dramatic illumination effects but the smaller size also works.  A 32" monitor provides about 100$^{\circ}$ horizontal field-of-view and about 70$^{\circ}$ vertical field-of-view. For the camera, we used an iPhone 8 mounted with a tripod behind the monitor for some of the experiments, and Samsung S10+  for others, demonstrating that our approach works with different types of cameras and monitors.  

We assume the monitor and camera are synchronized, e.g., by flashing an all-white frame at the start and at the end of the captured sequence, and setting both monitor and camera to operate at 30fps. We turn off auto focus and exposure and avoid any post-processing (e.g., night mode) that improves SNR but distorts the synchronous lighting and appearance information. The entire setup was easily replicated by another individual in their home.

\textbf{Evaluation Protocols} \label{sec:proto} To evaluate our approach we capture 19 sequences of 5 individuals watching YouTube videos, and use two protocols for evaluation. Individual A was captured  13 times with 7 different YouTube videos, and Individual B was captured 9 times with 8 different videos. All videos were captured at different days/times, with different ambient lighting and facial conditions. Individual C,D and E was captured once each, where D and E captured themselves in their own home using a 27" monitor and Samsung S10+ camera. Each captured sequence consists of a training video, followed by 2 test videos: 1) moving pattern on the monitor (to test directional lighting) and 2) a random YouTube clip.

The two evaluation protocols are:\\
\noindent Protocol 1 evaluates the typical light stage setting. At test time, we relight a pre-captured user sequence (which is part of the training) with a previously unseen lighting.\\
\noindent Protocol 2 evaluates generalization to unseen instances of a person and lighting conditions during test time. Given multiple captured sequences of a person over a period of time, we train relighting models and test it on a new user sequence with a previously unseen lighting. Note that the facial reflectance of the person during test time can vary from that of the training sequences. This protocol can be useful for relighting a person during video calls.

\textbf{Baselines} \label{sec:baseline} Potential baselines for our method are either performance relighting or portrait relighting methods.

Performance relighting typically assumes a linear relighting model combined with optical flow on specific frames captured in the light stage. Since we capture a person's appearance while watching a regular video, we do not have access to special frames viable for optical flow. The linear model as is (described in section  \ref{sec:ls}) fails to produce good results due to natural human motion, thus we omit those from the main paper (see appendix). 

Therefore, we focus on portrait relighting as our baseline with the caveat that such single-image methods were developed with a different application in mind. They are trained to work across many individuals, model lighting as a spherical environment map, and require only an image as input. In contrast we have access to input image and the source lighting, and we model lighting as a video frame.

\begin{table}[h]
\vspace{-0.5em}
\setlength\tabcolsep{5pt}
	\centering
	\small
	\captionsetup{justification=centering}
    	%\vspace{-.5em}
            \begin{tabular}{c|c|c|c}
            \hline
               & Linear model & RSun & Ours \\ 
                 \midrule
           Mean Absolute Error in \% & 3.37\% & 3.14\% & 3.35\% \\
                \hline
            \end{tabular} 
\vspace{-.5em}
		\caption{\small Error on mannequin sequence (no misalignment). Our implementation of RSun performs as expected in the ideal case.} 
	\label{tab:mann}
\vspace{-0.5em}
\end{table}

%%%%%%%%%% Table

\begin{table}
\vspace{-1em}
\setlength\tabcolsep{2pt}
	\centering
	\small
	\captionsetup{justification=centering}
    	%\vspace{-.5em}
            \begin{tabular}{r|c|c|c|}
            \hline
              Algorithm &  PSNR & RMSE & LPIPS\cite{zhang2018unreasonable}\\ 
               & (higher is better) & (lower is better) & (lower is better) \\
                 \midrule
           RSun & 24.51 &  0.0065 & 0.2718\\
           \hline
           Sun+ & 24.31 & 0.0067 & 0.1684 \\
           \hline
            Ours &  \textbf{25.21} & \textbf{0.0054 }& \textbf{0.1537} \\
                \hline
            \end{tabular} 
\vspace{-.5em}
		\caption{\small We report average PSNR, RMSE and LPIPS \cite{zhang2018unreasonable} score on 19 captured sequences. Protocol 1: Input image appears in training data, target lighting is unseen -- light stage.} 
	\label{tab:src2trg}
\vspace{-.5em}
\end{table}

\begin{table}
%\vspace{-1em}
\setlength\tabcolsep{2pt}
	\centering
	\small
	\captionsetup{justification=centering}
    	%\vspace{-.5em}
            \begin{tabular}{r|c|c|c|}
            \hline
              Algorithm &  PSNR & RMSE & LPIPS\cite{zhang2018unreasonable}\\ 
               & (higher is better) & (lower is better) & (lower is better) \\
                 \midrule
           Sun+ & 23.14 & 0.0064 & 0.1814 \\
           \hline
            Ours &  \textbf{26.36} & \textbf{0.0034}& \textbf{0.1417} \\
                \hline
            \end{tabular} 
\vspace{-.5em}
		\caption{\small We report average PSNR, RMSE and LPIPS \cite{zhang2018unreasonable} score on 8 captured sequences. Protocol 2: input image and target lighting is not part of the training data.}
	\label{tab:trg2trg}
\vspace{-.5em}
\end{table}

%%%%%%%%%%%%%%%%%%%%

Sun \etal \cite{sun2019single} is a state-of-the-art single image portrait relighting algorithm. It is trained on the OLAT (One Light At a Time) dataset \cite{sun2019single} consisting of light stage imagery of multiple individuals. Sun \etal's network consists of an U-Net which predicts both the relit image and the source lighting. It is trained on aligned pair of images with two different lighting conditions, generated from the OLAT data. It uses  L1 loss on relit image and L2 loss on predicted lighting, along with a reconstruction loss. 

Since neither the data nor model are available we re-implemented the network architecture introduced by Sun \etal and train it on the same training data as our algorithm. We use the exact same loss functions proposed in Sun \textit{et al.}. We call this implementation `RSun'. 
For perfectly aligned data (mannequin sequence), RSun produced good results, as expected (Table \ref{tab:mann}).  However, RSun performed worse on the rest of our datasets, due likely to the lack of perfect alignment of source and target photos (true for OLAT, but not true for our data). Our data has minor misalignment even after pose matching, which may be causing over-fitting in Sun \etal. We notice that the reconstruction loss is particularly problematic since during reconstruction path the input and the target is exactly aligned but they are not during the relighting path. 

Drawing inspirations from failures of RSun, we introduce a different set of loss functions that can handle the misalignment. We keep the L1 loss on the relit image and L2 loss over predicted lighting and remove the reconstruction loss. Then we add the perceptual loss eqn \ref{eq:p}, cycle consistency loss eqn \ref{eq:reg}, and adversarial loss with discriminator $D(\cdot)$, similar to our approach. This algorithm improves significantly over RSun in many examples. The difference between our approach and Sun+ lies in how lighting is infused into the network architecture. Comparison between RSun, Sun+ and our proposed approach helps us to understand how the loss functions and network architecture both contribute towards improving the quality of the relit results.

%%%%%%%%%%%% figures %%%%%%%%%%%%%

\begin{figure}[h!]
    \centering
    \newcolumntype{Y}{>{\centering\arraybackslash}X}
    \includegraphics[width=0.45\textwidth]{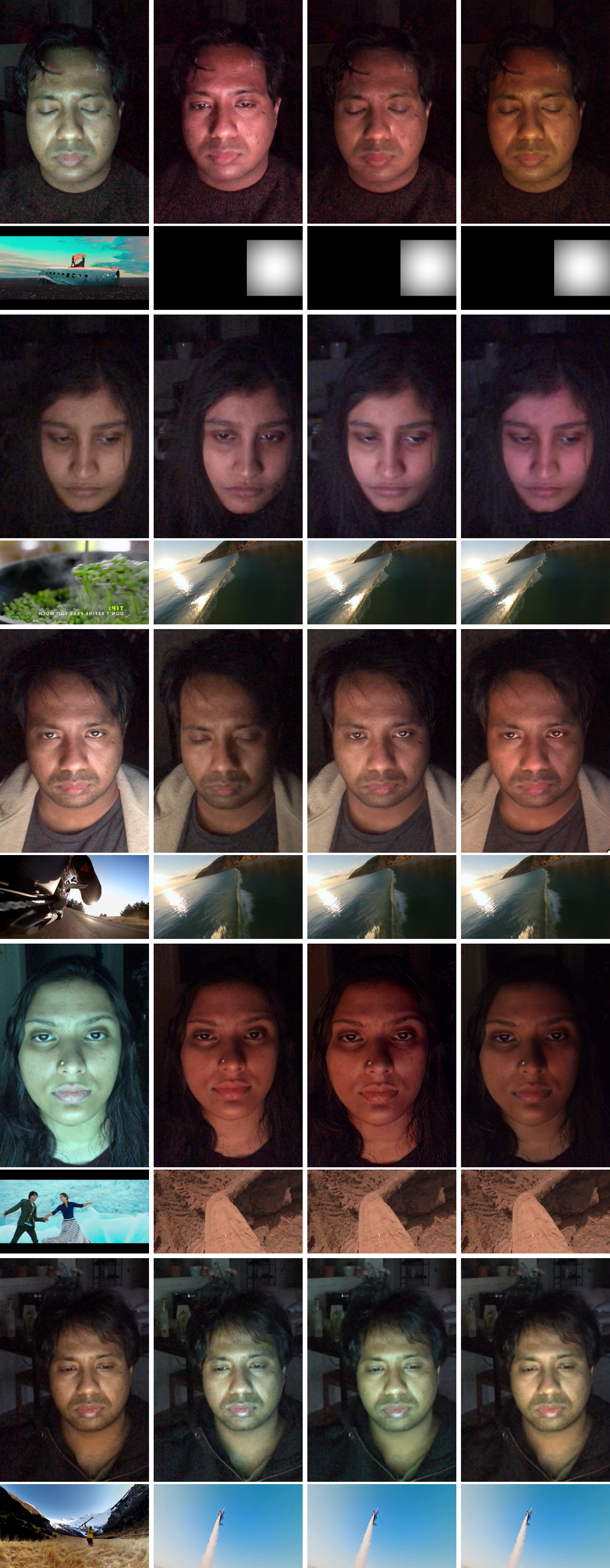}  %\vspace{-1.5em}
    \begin{footnotesize}
        \begin{tabularx}{0.45\textwidth}{YYYY}
          Input & Target & Ours & Sun+
        \end{tabularx}
    \end{footnotesize}
    \vspace{-0.5em}
    \caption{\small Qualitative comparison with Sun+ following Protocol 1 which resembles light stage testing setup, i.e., input image appears in the training data, while target lighting is unseen.} 
	\label{fig:src2trg}
    \vspace{-1.0em}
\end{figure}

\begin{figure}[h!]
    \centering
    \newcolumntype{Y}{>{\centering\arraybackslash}X}
    \includegraphics[width=0.45\textwidth]{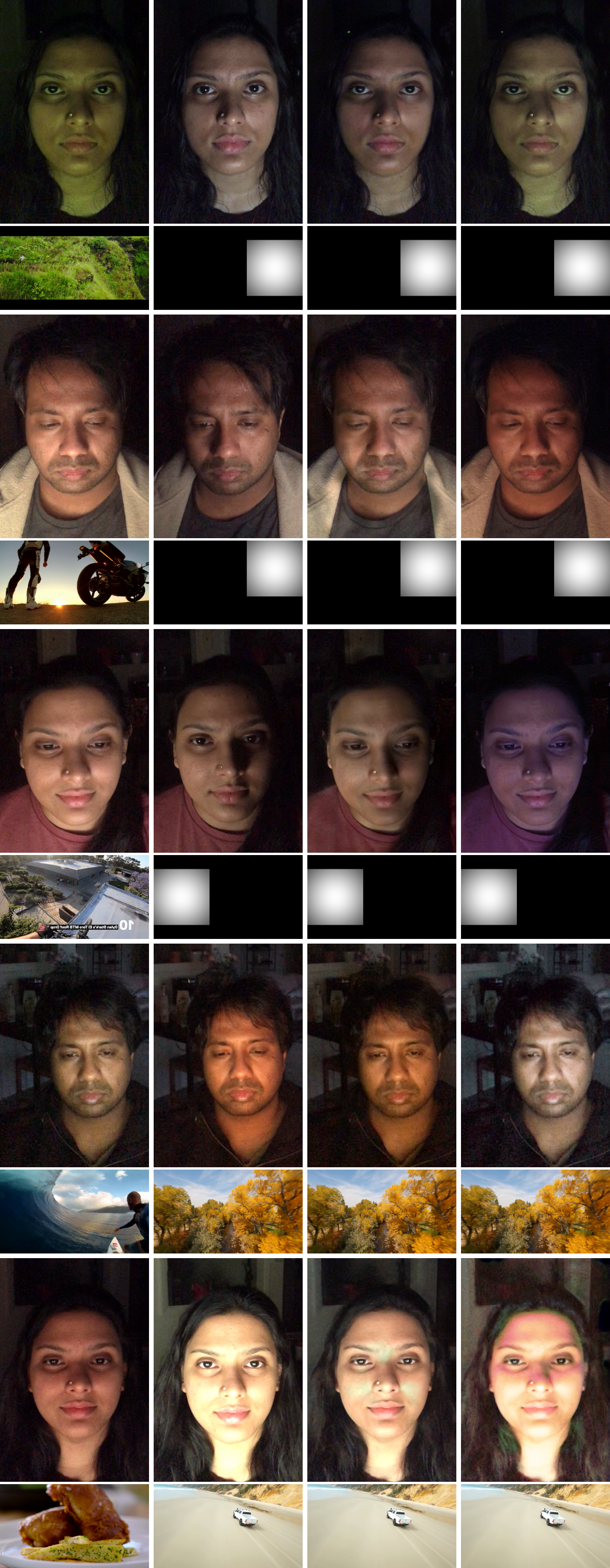}  %\vspace{-1.5em}
    \begin{footnotesize}
        \begin{tabularx}{0.45\textwidth}{YYYY}
          Input & Target & Ours & Sun+
        \end{tabularx}
    \end{footnotesize}
    \vspace{-0.5em}
    \caption{\small Qualitative comparison with Sun+ following Protocol 2 i.e input image and target lighting is not part of the training data.} 
	\label{fig:trg2trg}
    \vspace{-1.0em}
\end{figure}

\begin{figure*}[!ht]
	\centering
	\includegraphics[width=.95\textwidth]{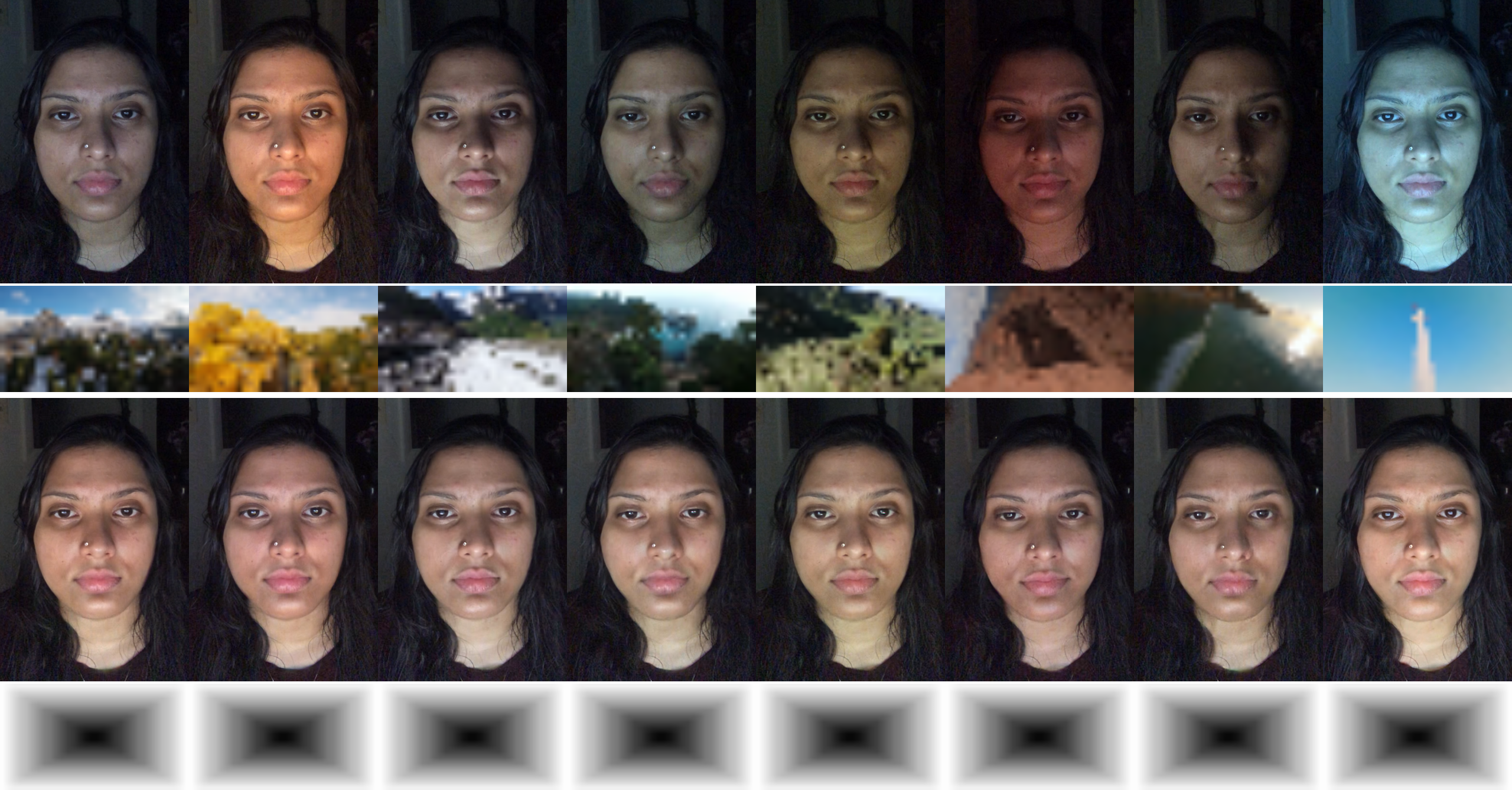}    \vspace{-.2cm}
	\caption{\small Improving lighting for video calls. During a video call in a poorly-lit room, lighting on the face may change with the content on the monitor (top row). We can re-lit the video with a ring-light pattern producing temporally consistent well-lit images (bottom row).}
	%\vspace{-1em}
	\label{fig:fixed}
\end{figure*}

\begin{figure*}[]
	\centering
	\includegraphics[width=.95\textwidth]{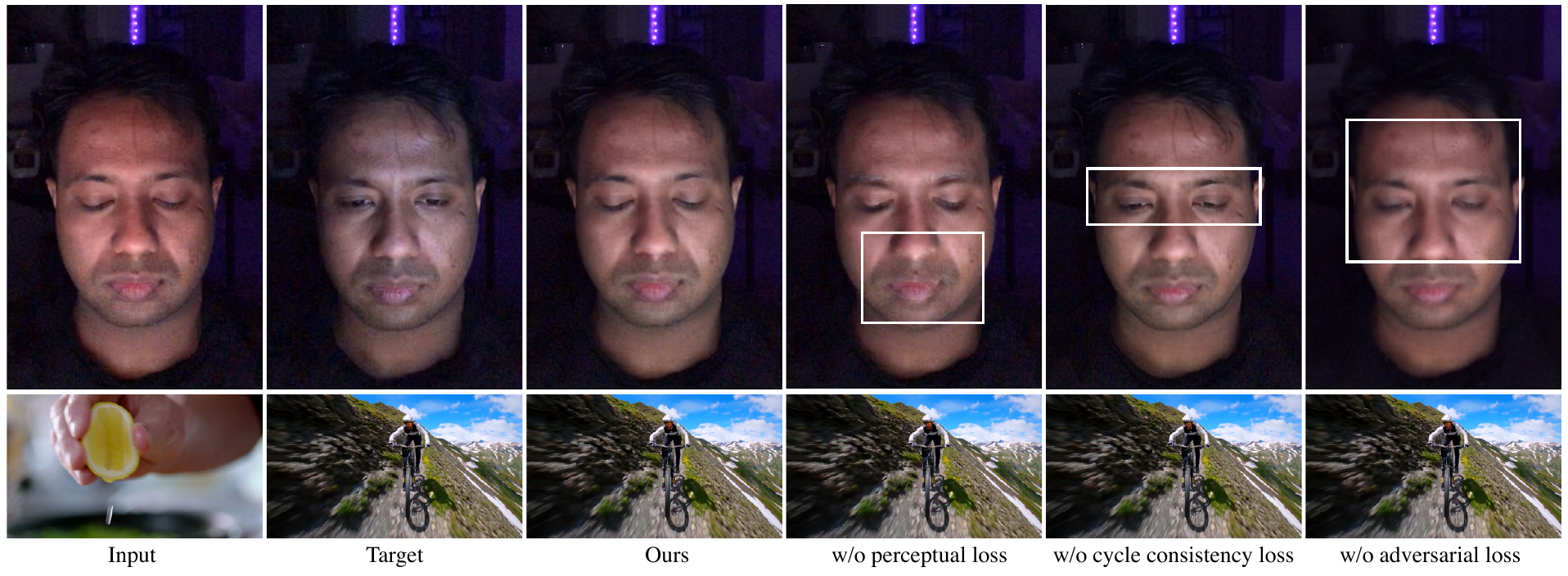}    \vspace{-.2cm}
	\caption{\small Ablation study w.r.t. loss functions. Both perceptual loss and adversarial loss helps in improving the quality of the relit images, while cycle consistency loss helps in maintaining the exact pose of the input image. }
	\vspace{-1em}
	\label{fig:abla_loss}
\end{figure*}

\begin{figure}[h!]
    \centering
    \newcolumntype{Y}{>{\centering\arraybackslash}X}
    \includegraphics[width=0.45\textwidth]{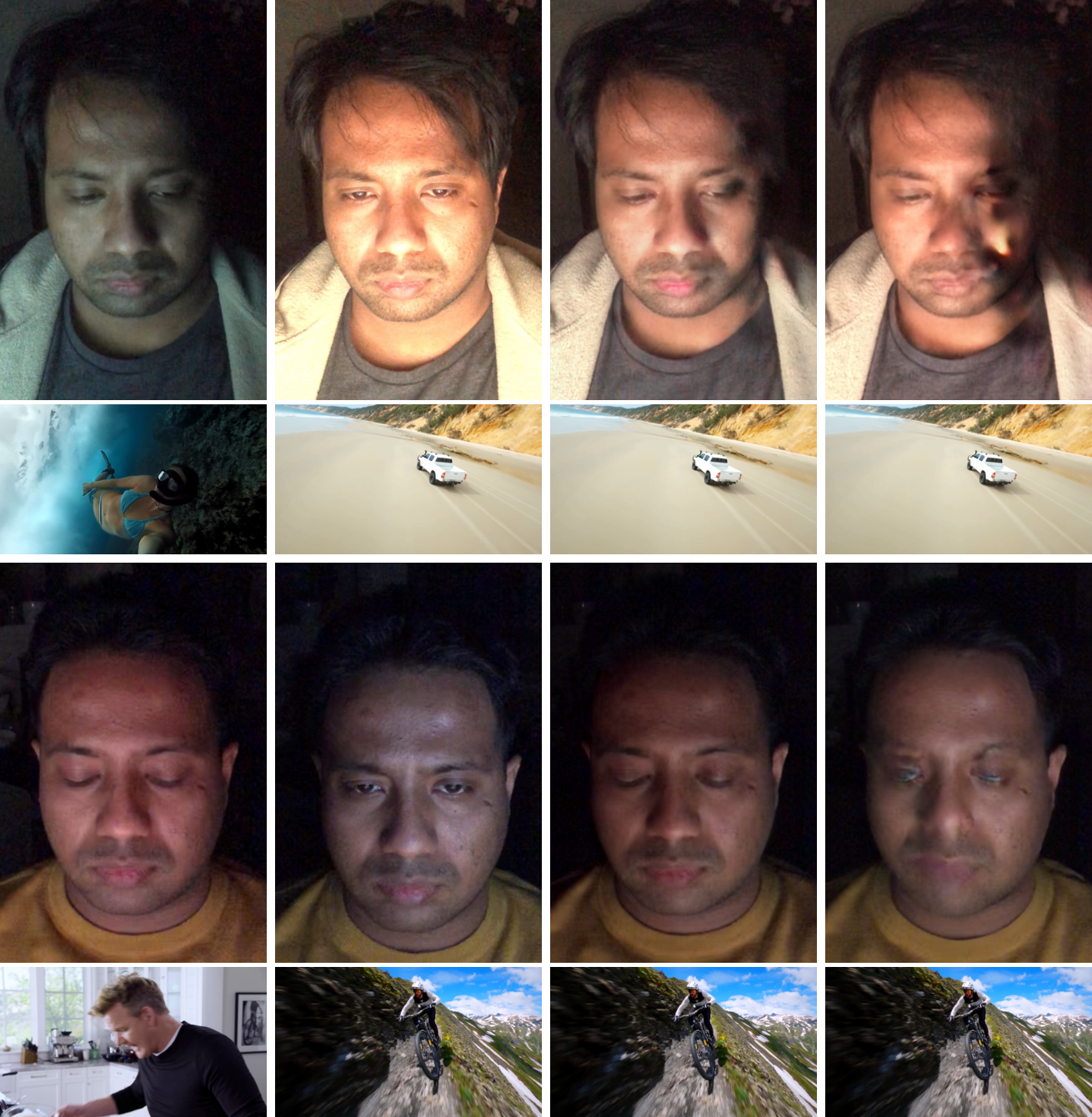}  %\vspace{-1.5em}
    \begin{footnotesize}
        \begin{tabularx}{0.45\textwidth}{YYYY}
          Input & Target & Ours & w/o source light
        \end{tabularx}
    \end{footnotesize}
    \vspace{-0.5em}
    \caption{\small Both the source and target monitor lighting is input to our network. Using source lighting improves the quality.} 
	\label{fig:abla_net}
   \vspace{-0.5em}
\end{figure}

\begin{figure}[h!]
    \centering
    \newcolumntype{Y}{>{\centering\arraybackslash}X}
    \includegraphics[width=0.45\textwidth]{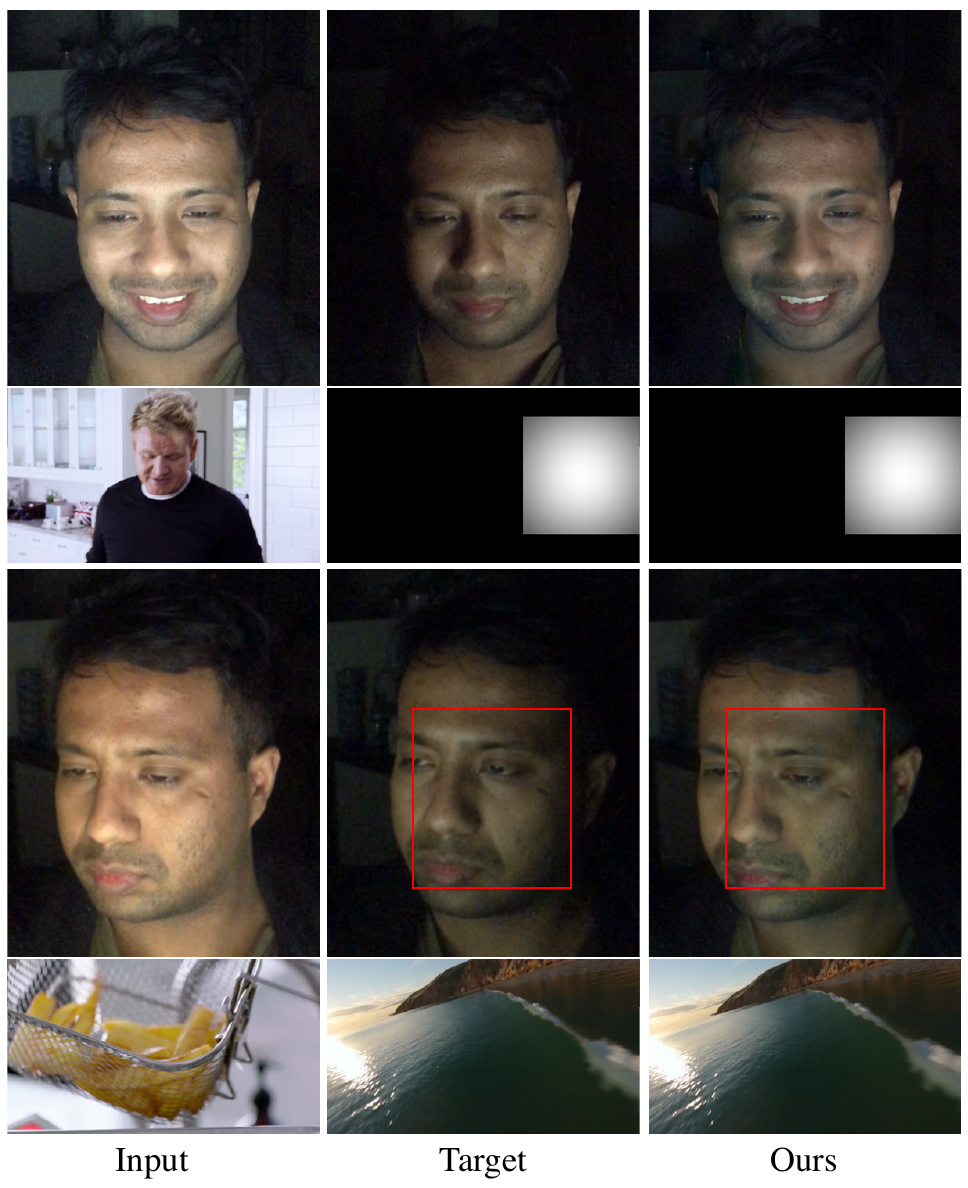}  %\vspace{-1.5em}
    \vspace{-0.5em}
    \caption{\small Although our relighting network is trained on nearly frontal faces with neutral expressions, it handle handle unseen expressions like a smiling face during test time (top row). However it is less accurate when the input image has a large pose variation from a frontal face (bottom row).} 
	\label{fig:robust}
    \vspace{-1.5em}
\end{figure}

%%%%%%%%%%%%%%%%%%%%%%%%%%%%%%%%

\subsection{Qualitative and Quantitative Comparison}

We present both qualitative and quantitative comparisons. Quantitative evaluation offers some challenges since we can't capture a person in exact same pose with 2 different lighting conditions. At test time, given the target lighting and the face captured with that lighting, we find the input face from the training sequence (in protocol 1), or test sequence (in protocol 2), which has the maximum overlap in face parsing, i.e., most similar pose. Even after finding the nearest pose match, the faces are not exactly the same, defeating measures like RMSE. Instead we use a perceptual measure, LPIPS \cite{zhang2018unreasonable}, to evaluate the quality of the relit images with that of the reference image captured under the same lighting. LPIPS also expects aligned images but is more robust to misalignment than RMSE. Since LPIPS with a VGG backbone is used as a loss function, AlexNet backbone is used for evaluation. 

\textbf{Protocol 1: Input image appears in training data, target light is unseen.} In Fig \ref{fig:src2trg} we present a qualitative comparison of our approach with that of RSun and Sun+. In Table \ref{tab:src2trg} we present perceptual similarity score LPIPS (lower is better), PSNR (higher is better) and RMSE (lower is better) on all 19 of our captured sequences. Both qualitative in Fig \ref{fig:src2trg} and quantitative evaluation in Table \ref{tab:src2trg} shows that our result is significantly better than RSun and Sun+.

\textbf{Protocol 2: Input image and target light are not part of training data.} Here we consider only individual A and B with 13 and 9 captured sequences each. We then create 4 evaluation setup for each of A and B by leaving 1 sequence out as the test set and considering the rest of the sequences as part of the training set. Test data is captured at different days/times compared to training data, with different YouTube videos. In Table \ref{tab:trg2trg} we present quantitative comparison with Sun+ by reporting LPIPS, RMSE and PSNR scores, qualitative evaluation is presented in  Fig \ref{fig:trg2trg}. We show that our approach can also perform relighting by changing light directions and color tones on input images unseen during the training. This means it can be deployed to perform real-time relighting, e.g., during a zoom video call.

\textbf{Improving lighting for video calls.} For the application of relighting a face during a video call, a ring-light pattern is a good choice to generate a well lit face.  In Fig \ref{fig:fixed} we show some image instances. We observe that the relit images remain well-lit and undergo minimal temporal change throughout the video. This follows Protocol 2, i.e. both the input image and lighting is unseen during the training.

\vspace{-0.5em}
\section{Ablation Study}
\label{sec:abla}
\vspace{-0.5em}
Our combination of data-processing, loss functions and neural architecture helps us to achieve high-quality relighting results from photos captured with monitor emitted illumination. Here we analyse how each component affects the quality of the result. We additionally train our network without using (a) adversarial loss with discriminator $D(\cdot)$, (b) perceptual loss in Eq.~\ref{eq:p}, and (c) cycle consistency loss in Eq.~\ref{eq:reg} separately. In Table~\ref{tab:abla} we compare the effect of each loss function on 3 captured sequences. Visual results in Figure \ref{fig:abla_loss} also shows that the results degrade in absence of perceptual and adversarial losses. On the other-hand in absence of cycle consistency loss, the relighting result changes the expression of the person -- eyes change from closed to open. Our network architecture takes both the source and the target monitor lighting as input. We show in Figure \ref{fig:abla_net} that knowing the source monitor lighting can improve the quality of the relit results.
\begin{table}[!h]
%\vspace{-0.5em}
\setlength\tabcolsep{5pt}
	\centering
	\small
	\captionsetup{justification=centering}
    	%\vspace{-.5em}
            \begin{tabular}{r|c|c|c|}
            \hline
               Algorithm &  LPIPS\cite{zhang2018unreasonable}\\ 
                & (lower is better) \\
                 \midrule
            Ours &  0.1311  \\
            \hline
             w/o perceptual loss & 0.1384 \\
             w/o cycle consistency loss & 0.1493 \\
            %\hline
            %\hline
              w/o adversarial loss & 0.1523 \\
             \hline
             w/o source light & 0.1374 \\
                \hline
            \end{tabular} 
\vspace{-.5em}
		\caption{\small Ablation study with perceptual similarity LPIPS \cite{zhang2018unreasonable}following Protocol 1 on 3 captured sequence, i.e., input image appears in the training data, while target lighting is unseen.}
	\label{tab:abla}
\vspace{-1em}
\end{table}

% \vspace{-1em}
\textbf{Robustness and Limitations} \label{sec:failure} Although our setup operates best in a dark room, we observe it can still handle moderate amount of background lighting as long as it is not directed towards  the subject. We empirically observe that on average many YouTube videos perform well as a source of lighting. Some categories of videos perform significantly better, e.g., videos with large camera motions, as they provide greater lighting variation. Conversely, videos with limited lighting variations, e.g., a single person talking or dark scenes, do not perform well.

Our method can also handle unseen expressions at test time. In Fig \ref{fig:robust} top row, we show an example of an individual smiling while the network was trained on only neutral expressions. However our method is far from perfect in presence of large pose variations in the input image. Cast shadows, while reproduced, can be lower-contrast compared to ground-truth, e.g., around the nose in Fig Fig \ref{fig:robust} bottom row. Application of our relighting network during video calls produces mostly  uniform and temporally stable results, but can exhibit flickering when the luminance of the input lighting drops significantly (see video results). Adding better alignment across faces, expression normalization, and lighting data augmentation might enable future improvements.

%\vspace{-0.5em}
\section{Conclusion}
\label{sec:discussion}
%\vspace{-0.5em}
This paper introduced a technique to democratize light stage capture for relighting applications, using nothing more than the monitor on your desk.  While our approach does not provide the full field of view, dynamic range, and full body coverage of a traditional laboratory light stage, it is easy to deploy and significantly expands access to light stage data and algorithms.  Furthermore, we show how relighting models can be trained passively, e.g., from footage of a user watching a normal movie.  Our relighting approach is robust to user motion, and produces realistic results.

\textbf{Ethics.} Our primary goal is to improve lighting for videos and video calls.  Our approach can also be used to improve privacy, by making it harder to infer screen content reflected from a user’s face.  We note, however, that synthetic image relighting is a form of image manipulation, and can facilitate compositing images for negative purposes as well.  Forgery detection and prevention is an important and ongoing topic of work.

\clearpage
\newpage

{\small
\bibliographystyle{ieee_fullname.bst}
\typeout{} 
\bibliography{ref.bib}
}

\clearpage
\renewcommand{\thesection}{\Alph{section}}
\setcounter{section}{0}

\section{Appendix}
\label{sec:append}

\subsection{Overview}
\begin{figure}
	\centering
	\includegraphics[width=.3\textwidth]{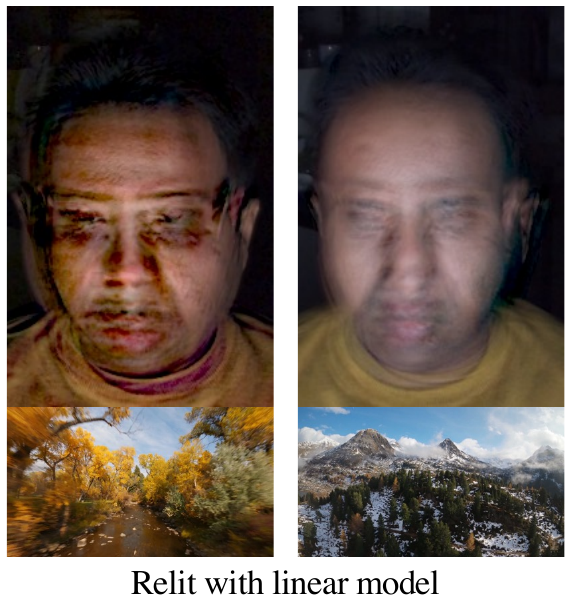}    \vspace{-.2cm}
	\caption{\small Linear model fails on humans captured with natural head motion.} 
	\vspace{-1em}
	\label{fig:lin_fail}
\end{figure}

\begin{figure}
	\centering
	\includegraphics[width=.45\textwidth]{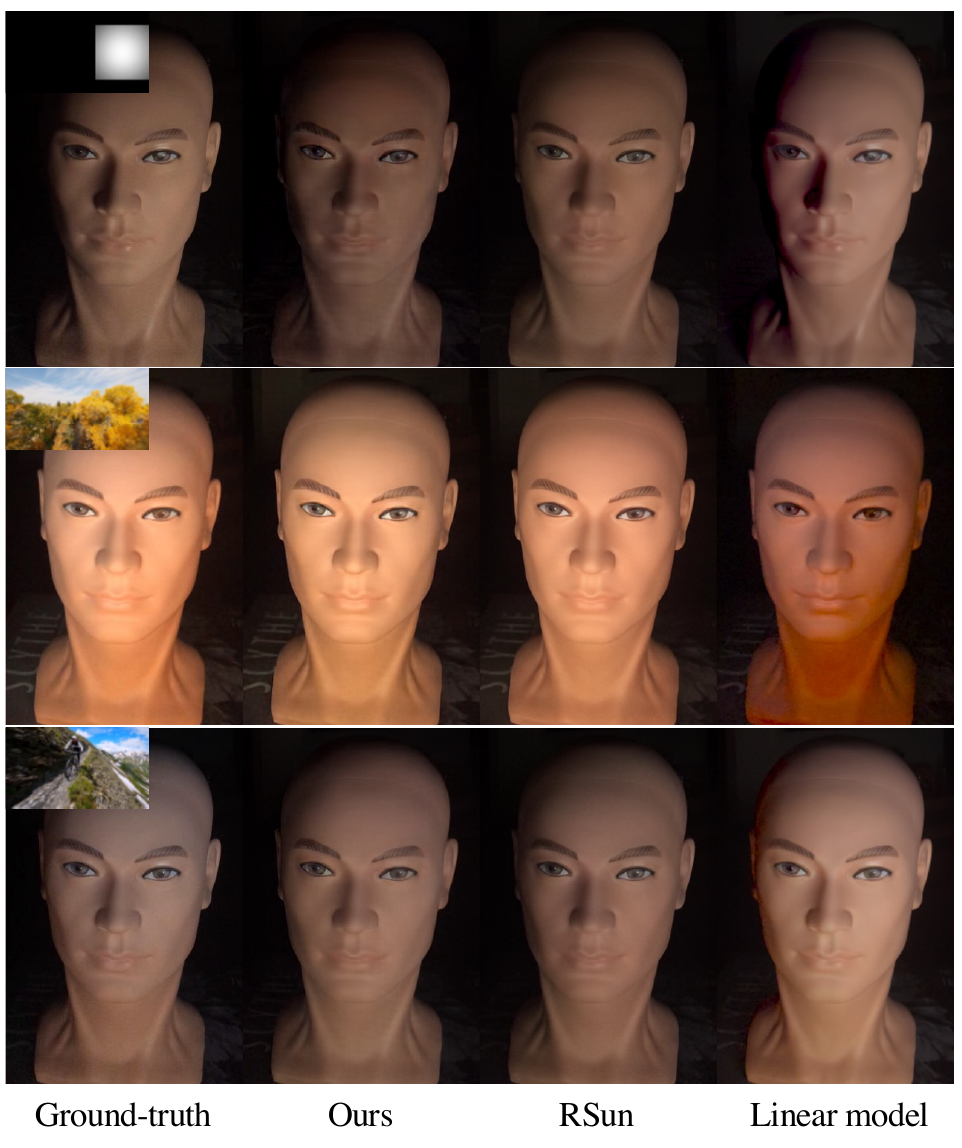}    \vspace{-.2cm}
	\caption{\small Qualitative Comparison on mannequin sequence. Our method predicts better color with natural images from the test-clip sequence.} 
	\vspace{-1em}
	\label{fig:man}
\end{figure}

\begin{figure*}[h!]
    \centering
    \newcolumntype{A}{>{\hsize=0.202\textwidth}X}
    \newcolumntype{B}{>{\hsize=0.086\textwidth}X}
    \newcolumntype{Y}{>{\centering\arraybackslash}X}
    \includegraphics[width=0.6\textwidth]{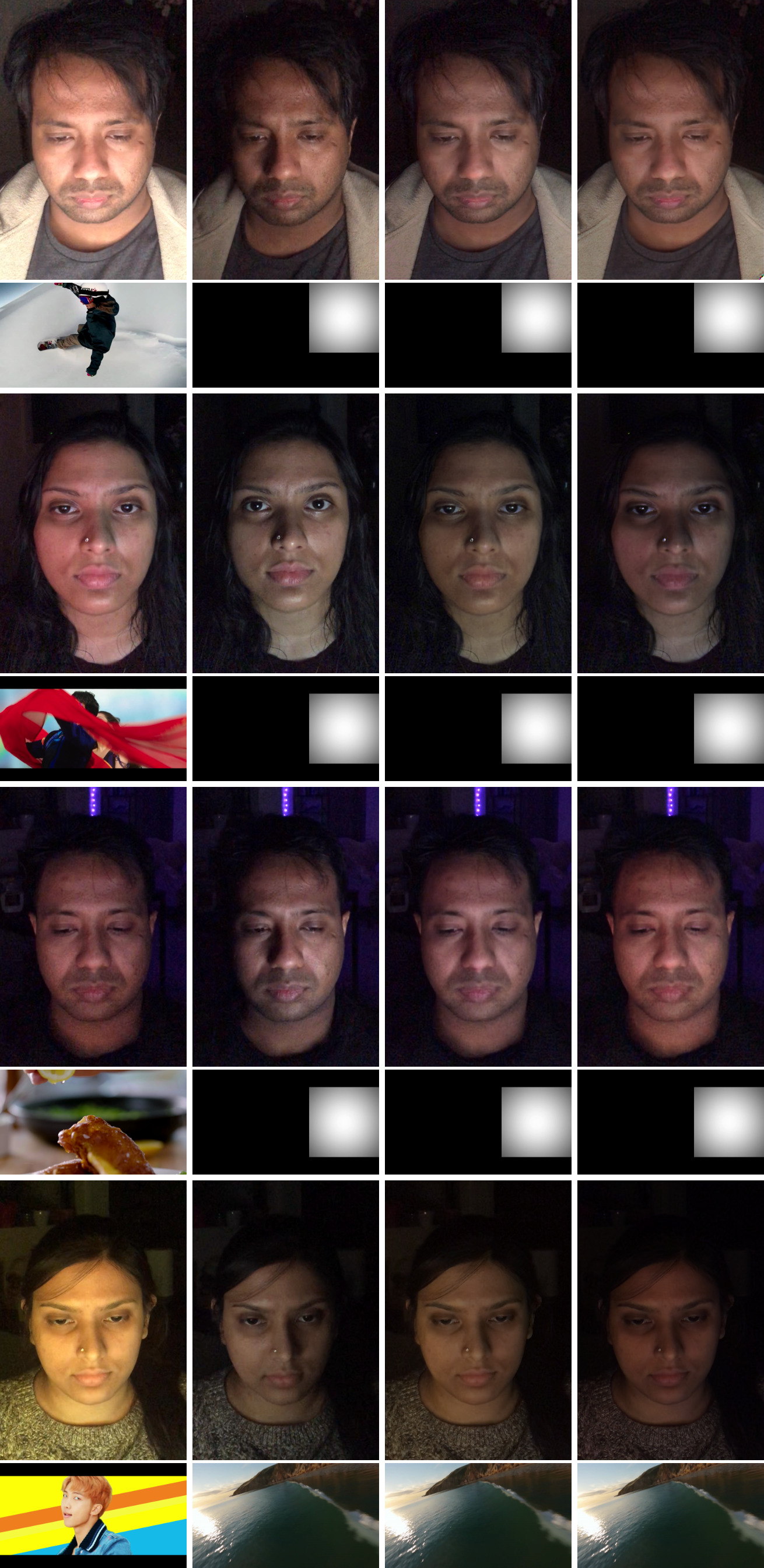}
    \begin{footnotesize}
        \begin{tabularx}{0.6\textwidth}{YYYY}
          Input & Target & Ours & Sun+
        \end{tabularx}
    \end{footnotesize}
    \vspace{-0.5em}
    \caption{\small Qualitative comparison with Sun+ following Protocol 1 which resembles light stage testing setup, i.e., input image appears in the training data, while target lighting is unseen.}
    \label{fig:src2trg1}
    \vspace{-1.0em}
\end{figure*}

\begin{figure*}[h!]
    \centering
    \newcolumntype{A}{>{\hsize=0.202\textwidth}X}
    \newcolumntype{B}{>{\hsize=0.086\textwidth}X}
    \newcolumntype{Y}{>{\centering\arraybackslash}X}
    \includegraphics[width=0.6\textwidth]{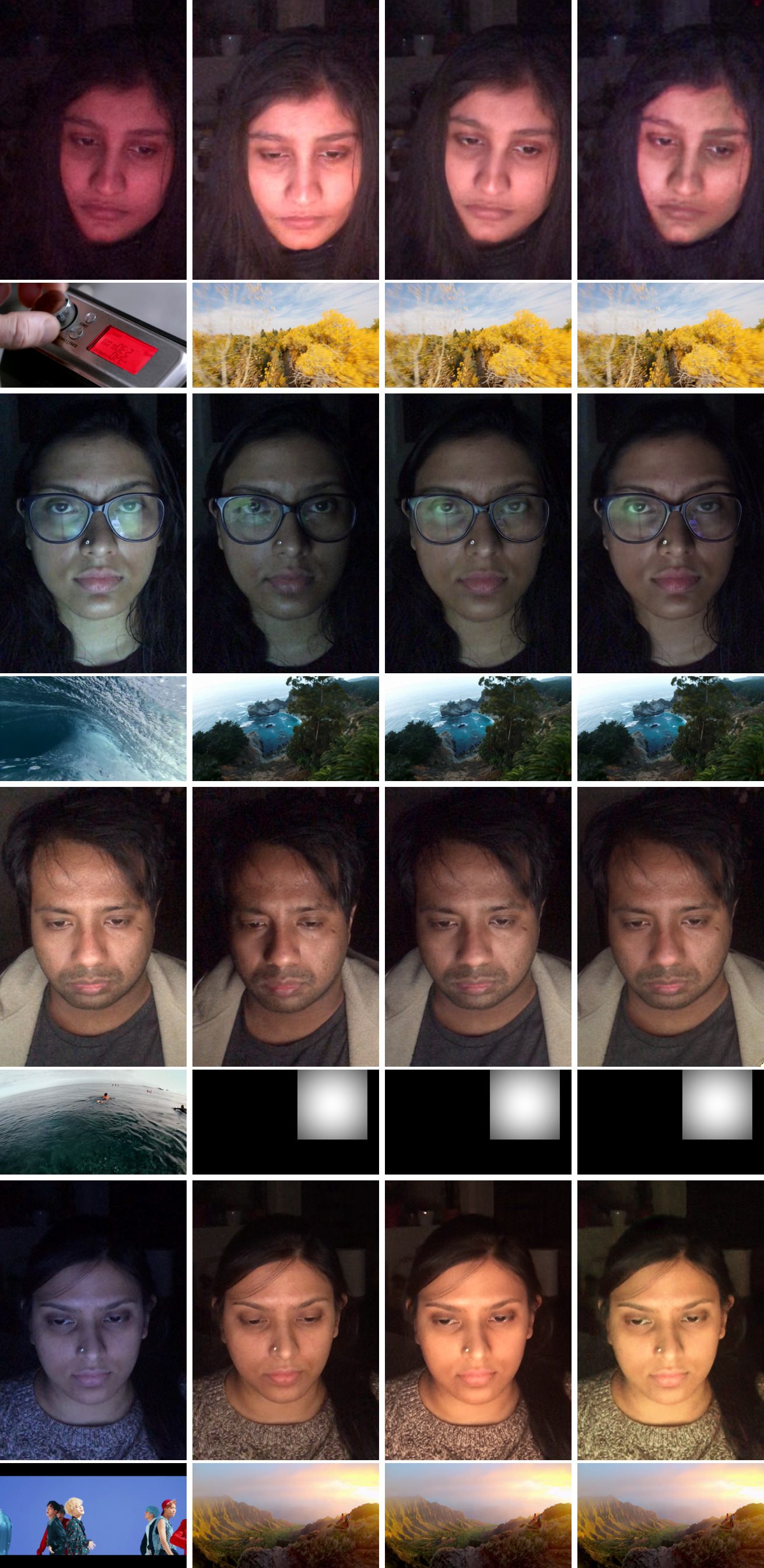}
    \begin{footnotesize}
        \begin{tabularx}{0.6\textwidth}{YYYY}
          Input & Target & Ours & Sun+
        \end{tabularx}
    \end{footnotesize}
    \vspace{-0.5em}
    \caption{\small Qualitative comparison with Sun+ following Protocol 1 which resembles light stage testing setup, i.e., input image appears in the training data, while target lighting is unseen.}
    \label{fig:src2trg2}
    \vspace{-1.0em}
\end{figure*}

\begin{figure*}[h!]
    \centering
    \newcolumntype{A}{>{\hsize=0.202\textwidth}X}
    \newcolumntype{B}{>{\hsize=0.086\textwidth}X}
    \newcolumntype{Y}{>{\centering\arraybackslash}X}
    \includegraphics[width=0.6\textwidth]{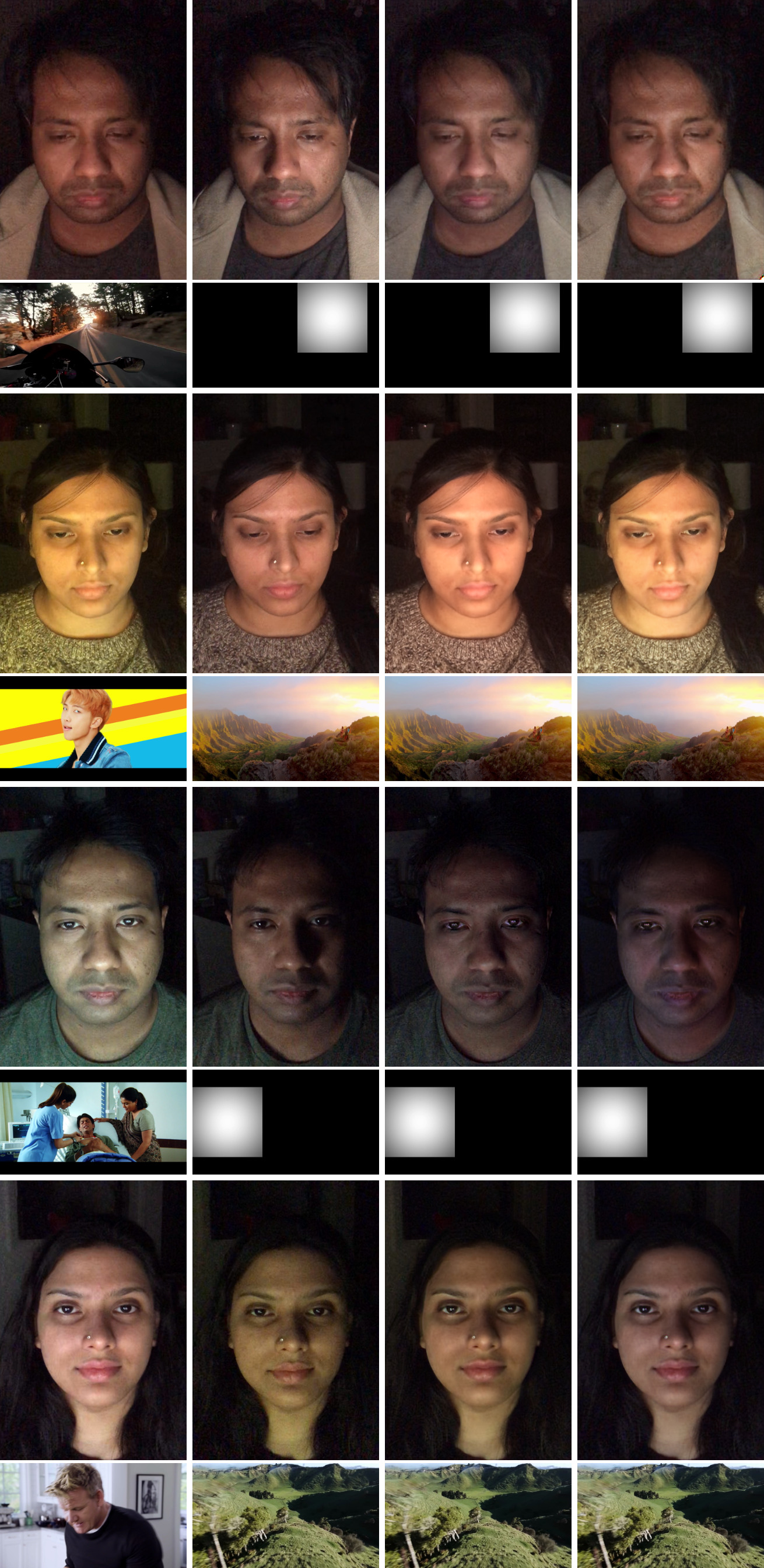}
    \begin{footnotesize}
        \begin{tabularx}{0.6\textwidth}{YYYY}
          Input & Target & Ours & Sun+
        \end{tabularx}
    \end{footnotesize}
    \vspace{-0.5em}
    \caption{\small Qualitative comparison with Sun+ following Protocol 1 which resembles light stage testing setup, i.e., input image appears in the training data, while target lighting is unseen.}
    \label{fig:src2trg3}
    \vspace{-1.0em}
\end{figure*}

\begin{figure*}[h!]
    \centering
    \newcolumntype{A}{>{\hsize=0.202\textwidth}X}
    \newcolumntype{B}{>{\hsize=0.086\textwidth}X}
    \newcolumntype{Y}{>{\centering\arraybackslash}X}
    \includegraphics[width=0.6\textwidth]{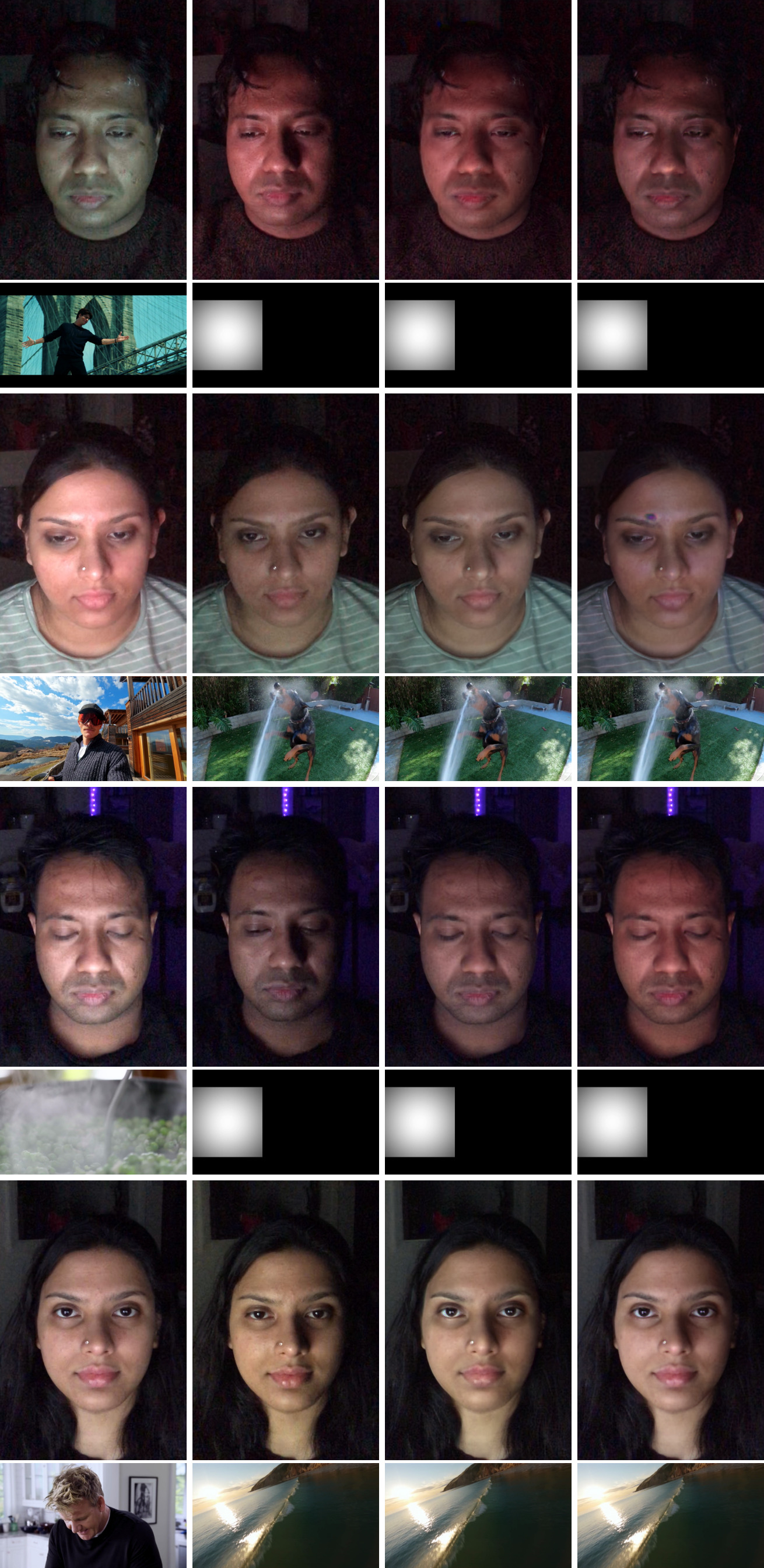}
    \begin{footnotesize}
        \begin{tabularx}{0.6\textwidth}{YYYY}
          Input & Target & Ours & Sun+
        \end{tabularx}
    \end{footnotesize}
    \vspace{-0.5em}
    \caption{\small Qualitative comparison with Sun+ following Protocol 1 which resembles light stage testing setup, i.e., input image appears in the training data, while target lighting is unseen.}
    \label{fig:src2trg4}
    \vspace{-1.0em}
\end{figure*}

\begin{figure*}[h!]
    \centering
    \newcolumntype{A}{>{\hsize=0.202\textwidth}X}
    \newcolumntype{B}{>{\hsize=0.086\textwidth}X}
    \newcolumntype{Y}{>{\centering\arraybackslash}X}
    \includegraphics[width=0.6\textwidth]{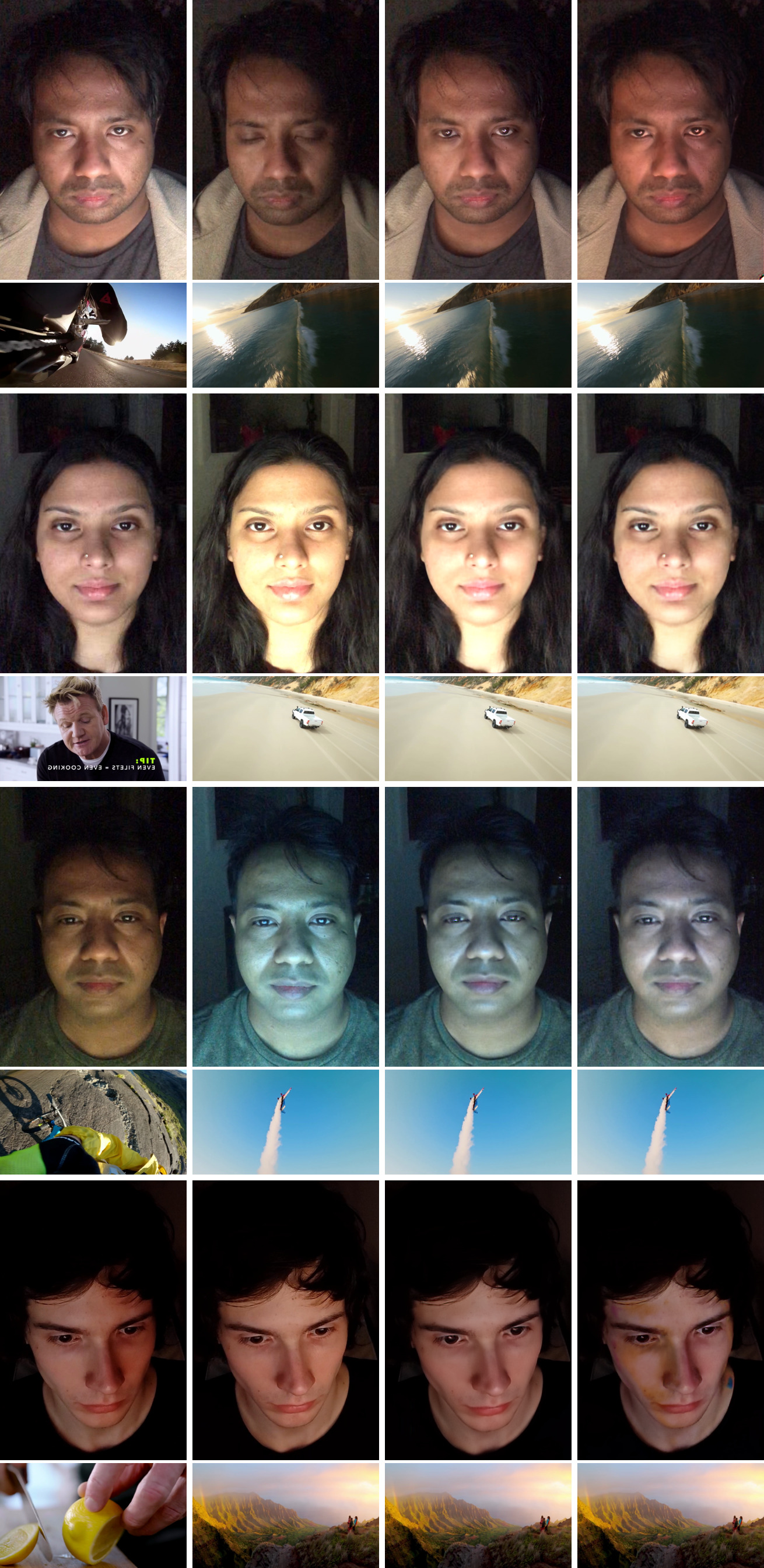}
    \begin{footnotesize}
        \begin{tabularx}{0.6\textwidth}{YYYY}
          Input & Target & Ours & Sun+
        \end{tabularx}
    \end{footnotesize}
    \vspace{-0.5em}
    \caption{\small Qualitative comparison with Sun+ following Protocol 1 which resembles light stage testing setup, i.e., input image appears in the training data, while target lighting is unseen.}
    \label{fig:src2trg5}
    \vspace{-1.0em}
\end{figure*}

\begin{figure*}[h!]
    \centering
    \newcolumntype{A}{>{\hsize=0.202\textwidth}X}
    \newcolumntype{B}{>{\hsize=0.086\textwidth}X}
    \newcolumntype{Y}{>{\centering\arraybackslash}X}
    \includegraphics[width=0.6\textwidth]{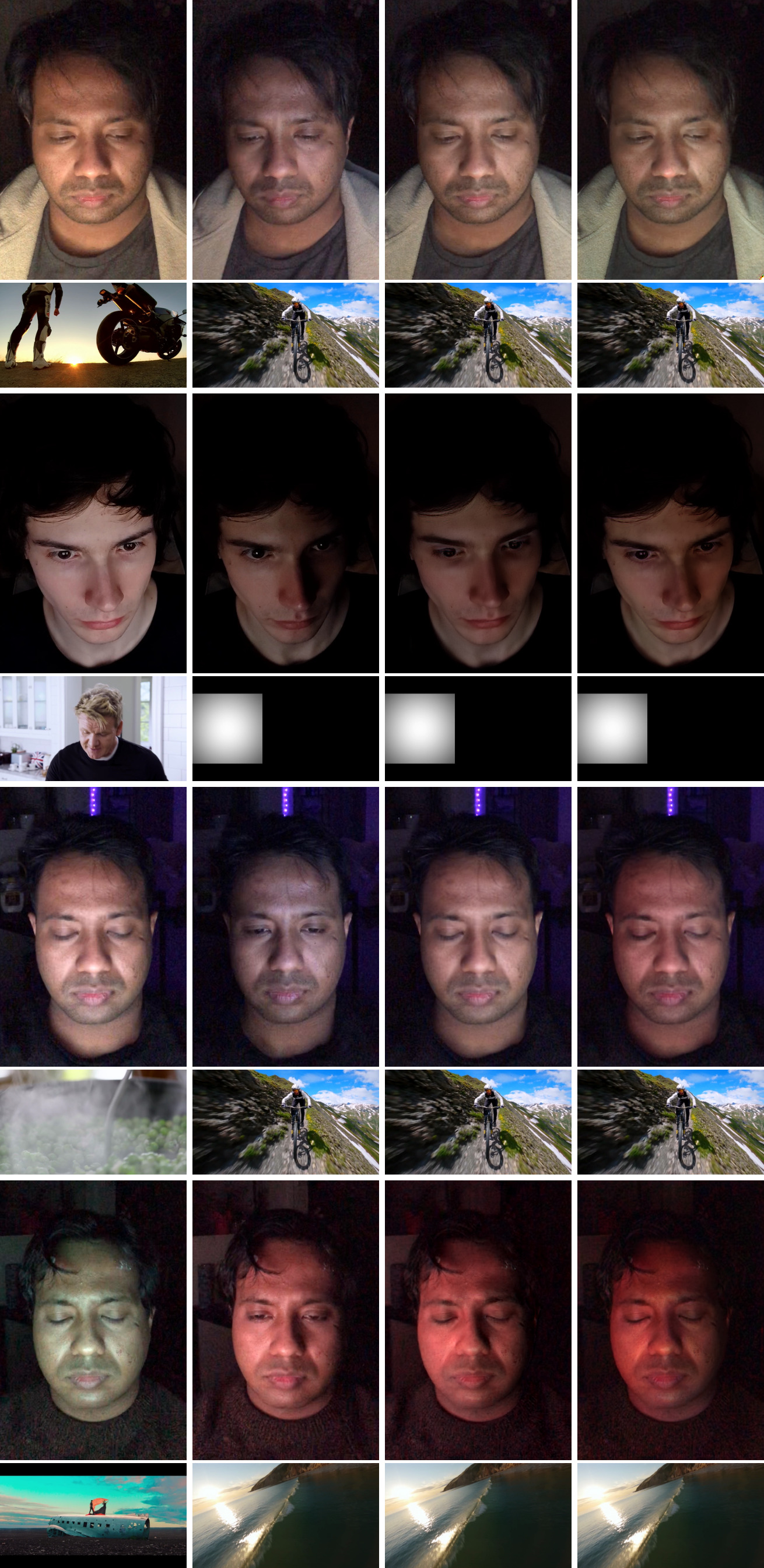}
    \begin{footnotesize}
        \begin{tabularx}{0.6\textwidth}{YYYY}
          Input & Target & Ours & Sun+
        \end{tabularx}
    \end{footnotesize}
    \vspace{-0.5em}
    \caption{\small Qualitative comparison with Sun+ following Protocol 1 which resembles light stage testing setup, i.e., input image appears in the training data, while target lighting is unseen.}
    \label{fig:src2trg6}
    \vspace{-1.0em}
\end{figure*}

\begin{figure*}[h!]
    \centering
    \newcolumntype{A}{>{\hsize=0.202\textwidth}X}
    \newcolumntype{B}{>{\hsize=0.086\textwidth}X}
    \newcolumntype{Y}{>{\centering\arraybackslash}X}
    \includegraphics[width=0.6\textwidth]{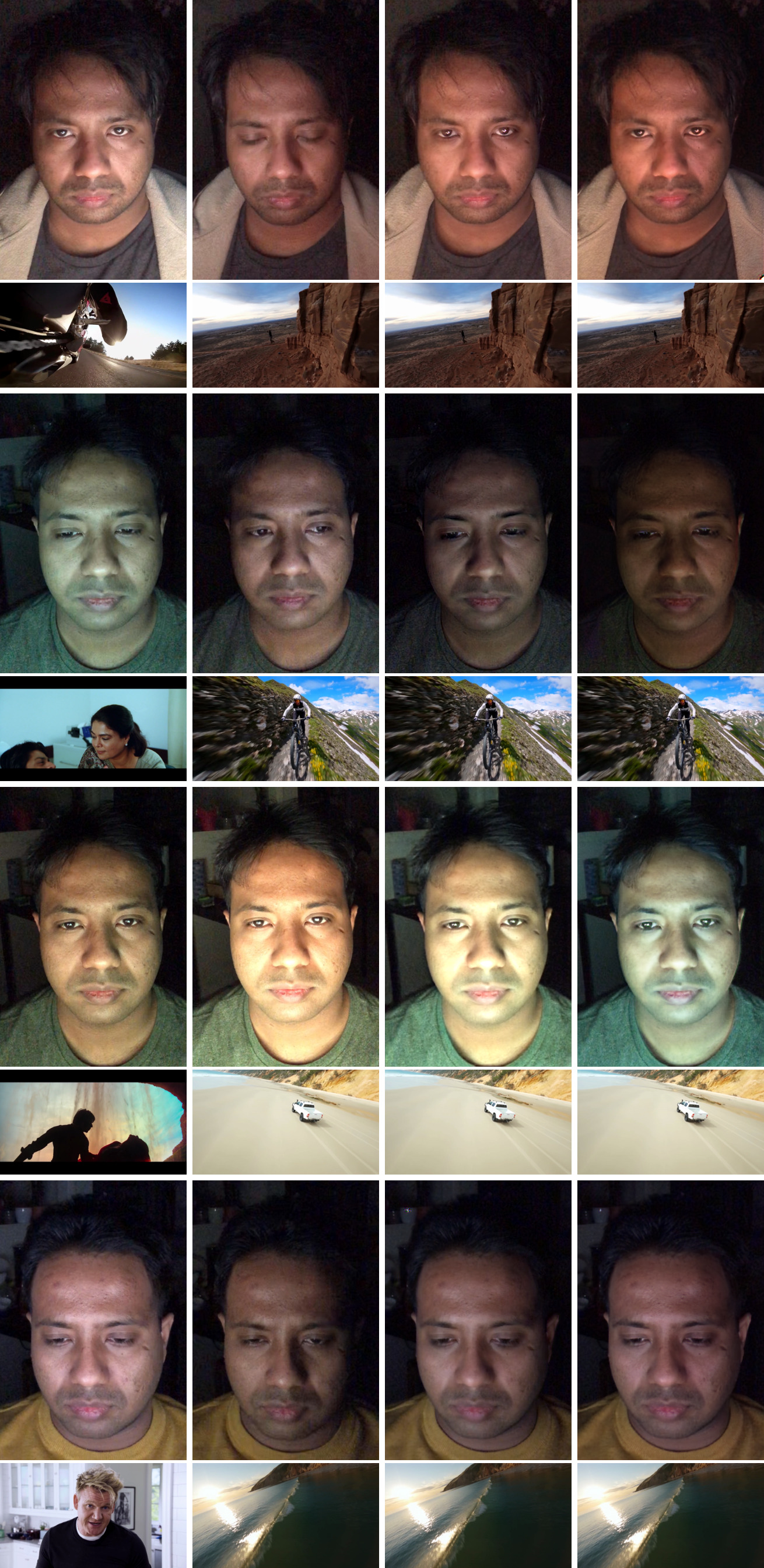}
    \begin{footnotesize}
        \begin{tabularx}{0.6\textwidth}{YYYY}
          Input & Target & Ours & Sun+
        \end{tabularx}
    \end{footnotesize}
    \vspace{-0.5em}
    \caption{\small Qualitative comparison with Sun+ following Protocol 1 which resembles light stage testing setup, i.e., input image appears in the training data, while target lighting is unseen.}
    \label{fig:src2trg7}
    \vspace{-1.0em}
\end{figure*}

\begin{figure*}[h!]
    \centering
    \newcolumntype{A}{>{\hsize=0.202\textwidth}X}
    \newcolumntype{B}{>{\hsize=0.086\textwidth}X}
    \newcolumntype{Y}{>{\centering\arraybackslash}X}
    \includegraphics[width=0.6\textwidth]{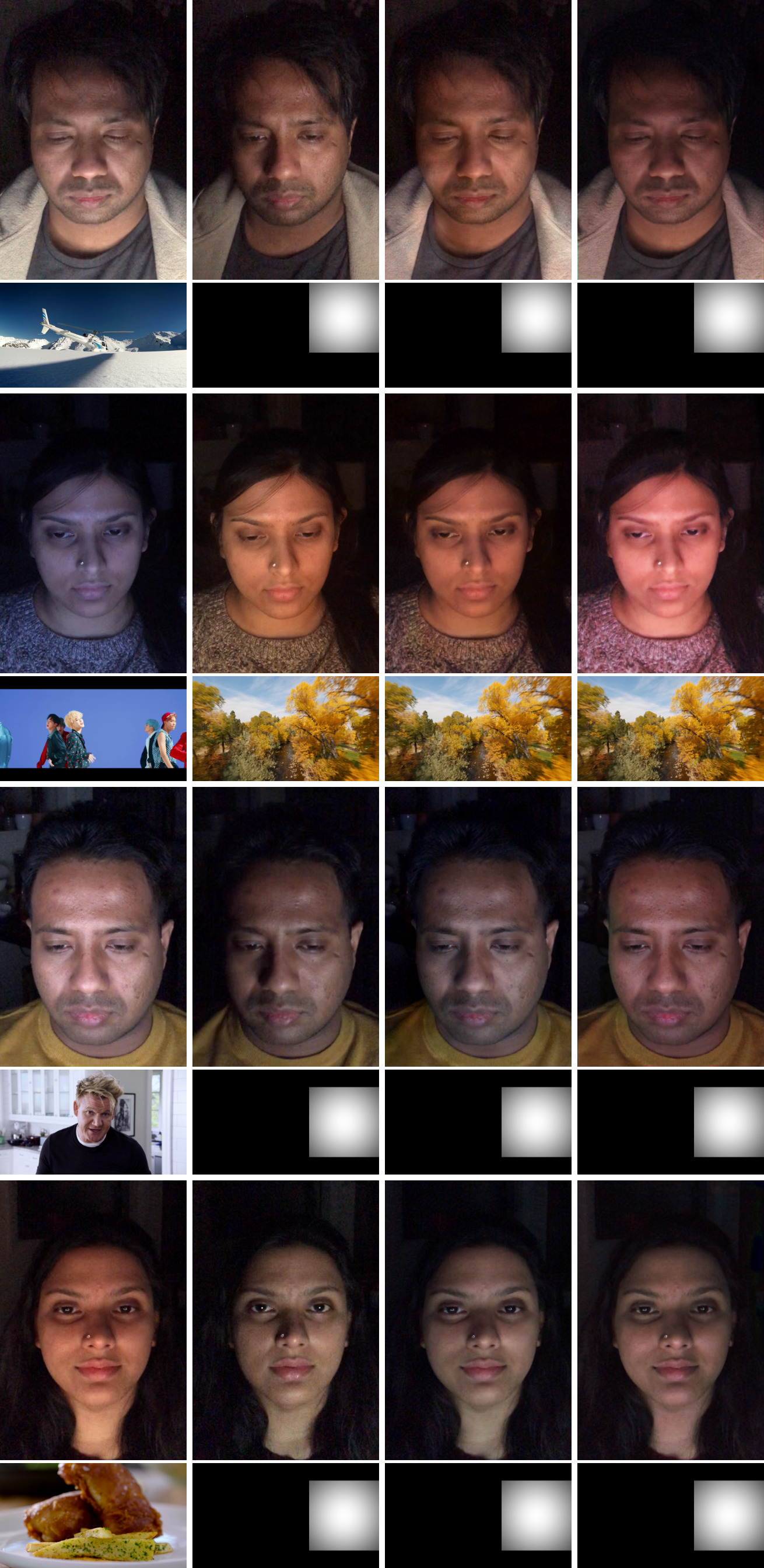}
    \begin{footnotesize}
        \begin{tabularx}{0.6\textwidth}{YYYY}
          Input & Target & Ours & Sun+
        \end{tabularx}
    \end{footnotesize}
    \vspace{-0.5em}
    \caption{\small Qualitative comparison with  Sun+ following Protocol 2 i.e input image and target relighting is not part of the training data.}
    \label{fig:trg2trg1}
    \vspace{-1.0em}
\end{figure*}

\begin{figure*}[h!]
    \centering
    \newcolumntype{A}{>{\hsize=0.202\textwidth}X}
    \newcolumntype{B}{>{\hsize=0.086\textwidth}X}
    \newcolumntype{Y}{>{\centering\arraybackslash}X}
    \includegraphics[width=0.6\textwidth]{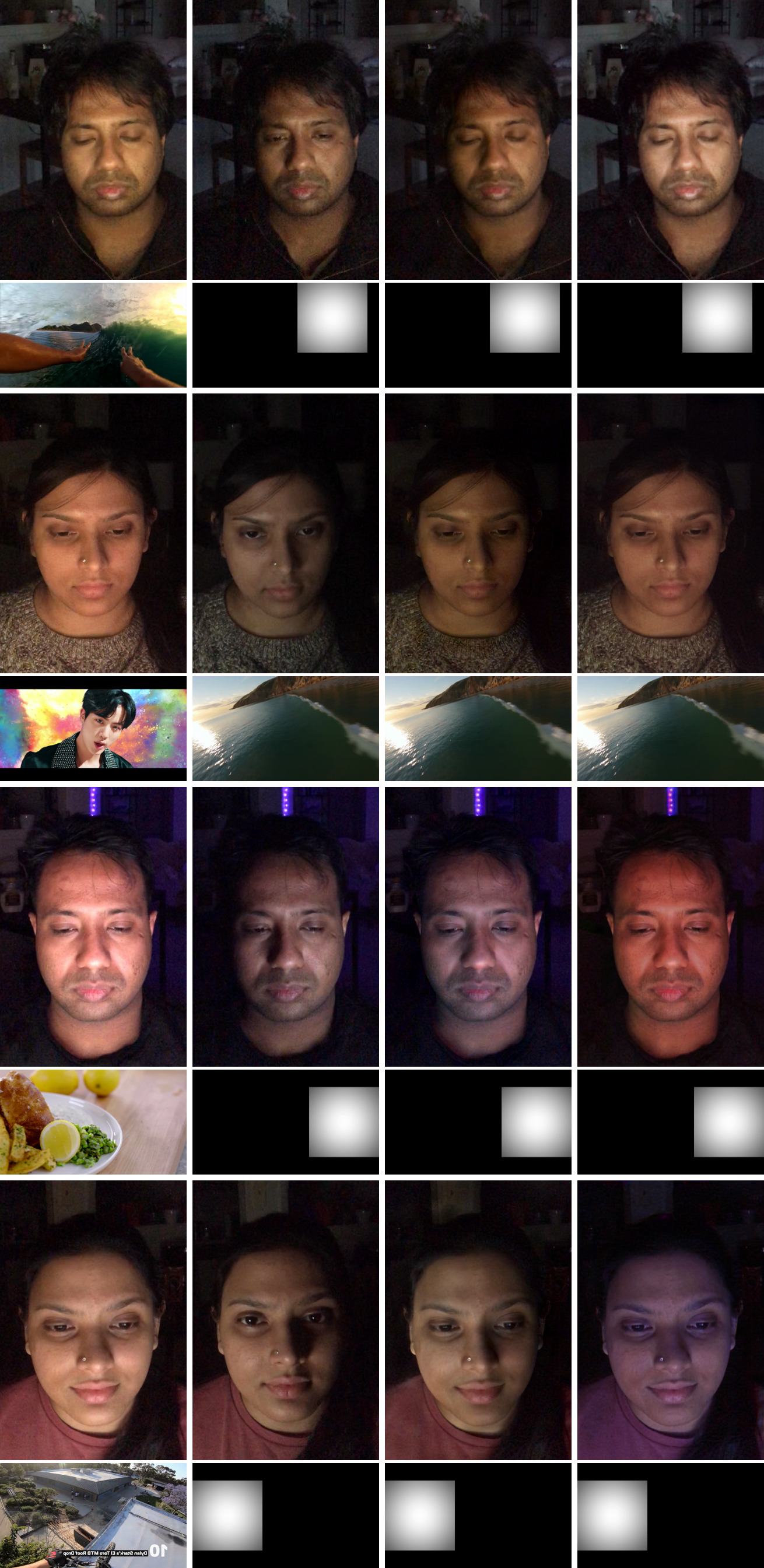}
    \begin{footnotesize}
        \begin{tabularx}{0.6\textwidth}{YYYY}
          Input & Target & Ours & Sun+
        \end{tabularx}
    \end{footnotesize}
    \vspace{-0.5em}
    \caption{\small Qualitative comparison with  Sun+ following Protocol 2 i.e input image and target relighting is not part of the training data.}
    \label{fig:trg2trg2}
    \vspace{-1.0em}
\end{figure*}

\begin{figure*}[h!]
    \centering
    \newcolumntype{A}{>{\hsize=0.202\textwidth}X}
    \newcolumntype{B}{>{\hsize=0.086\textwidth}X}
    \newcolumntype{Y}{>{\centering\arraybackslash}X}
    \includegraphics[width=0.6\textwidth]{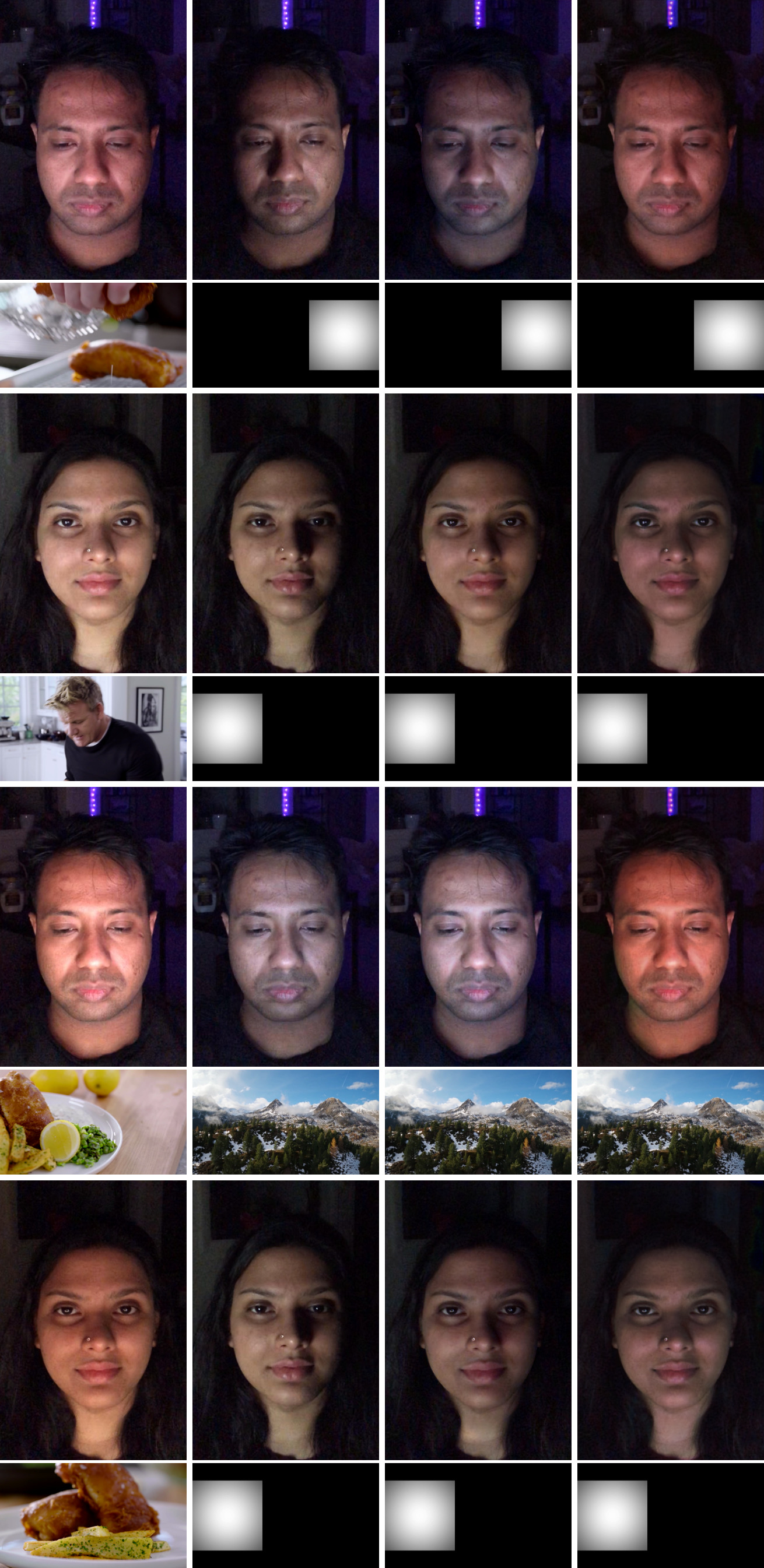}
    \begin{footnotesize}
        \begin{tabularx}{0.6\textwidth}{YYYY}
          Input & Target & Ours & Sun+
        \end{tabularx}
    \end{footnotesize}
    \vspace{-0.5em}
    \caption{\small Qualitative comparison with  Sun+ following Protocol 2 i.e input image and target relighting is not part of the training data.}
    \label{fig:trg2trg3}
    \vspace{-1.0em}
\end{figure*}

\begin{figure*}[h!]
    \centering
    \newcolumntype{A}{>{\hsize=0.202\textwidth}X}
    \newcolumntype{B}{>{\hsize=0.086\textwidth}X}
    \newcolumntype{Y}{>{\centering\arraybackslash}X}
    \includegraphics[width=0.6\textwidth]{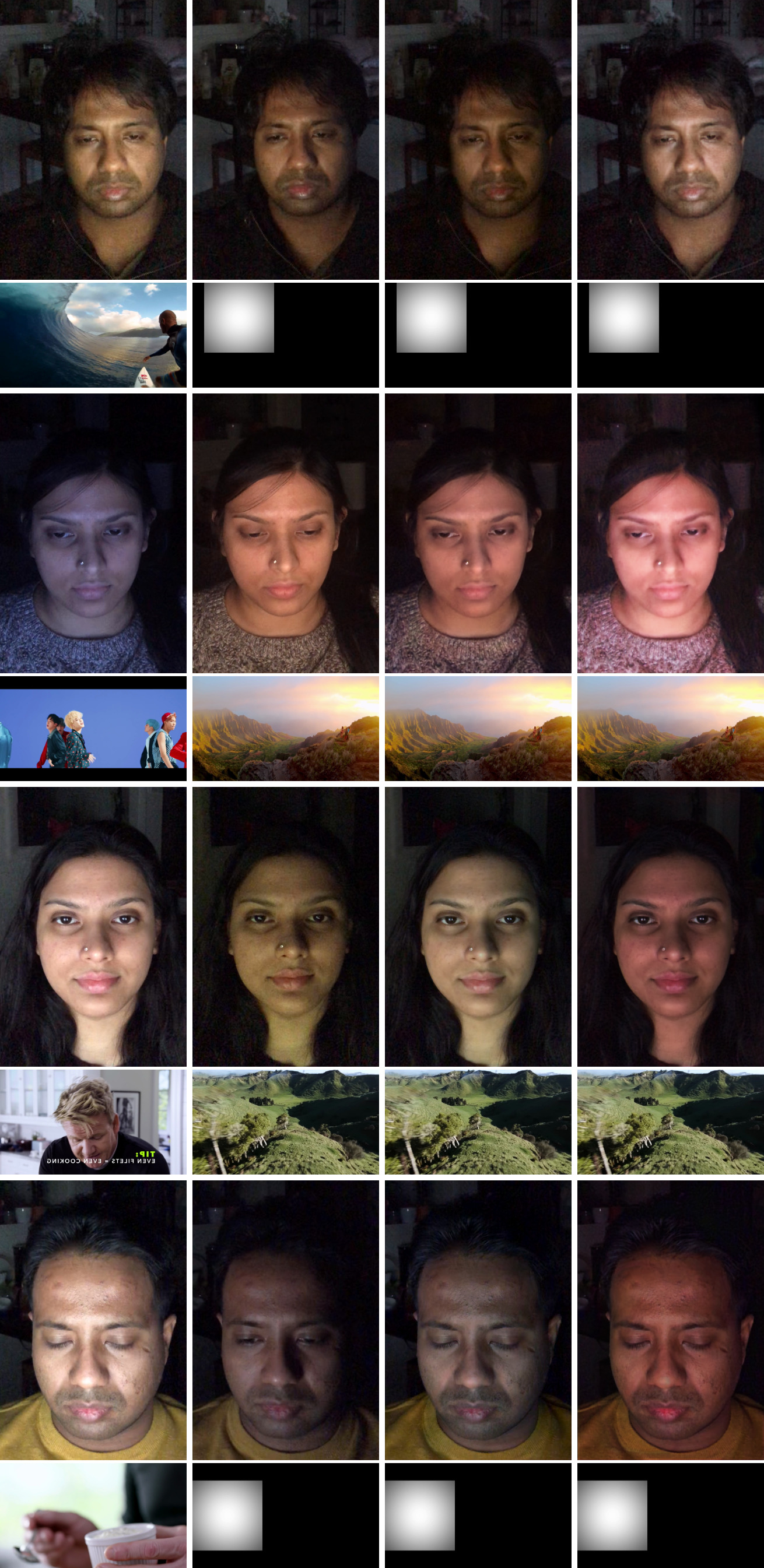}
    \begin{footnotesize}
        \begin{tabularx}{0.6\textwidth}{YYYY}
          Input & Target & Ours & Sun+
        \end{tabularx}
    \end{footnotesize}
    \vspace{-0.5em}
    \caption{\small Qualitative comparison with  Sun+ following Protocol 2 i.e input image and target relighting is not part of the training data.}
    \label{fig:trg2trg4}
    \vspace{-1.0em}
\end{figure*}

\begin{figure*}[h!]
    \centering
    \newcolumntype{A}{>{\hsize=0.202\textwidth}X}
    \newcolumntype{B}{>{\hsize=0.086\textwidth}X}
    \newcolumntype{Y}{>{\centering\arraybackslash}X}
    \includegraphics[width=0.6\textwidth]{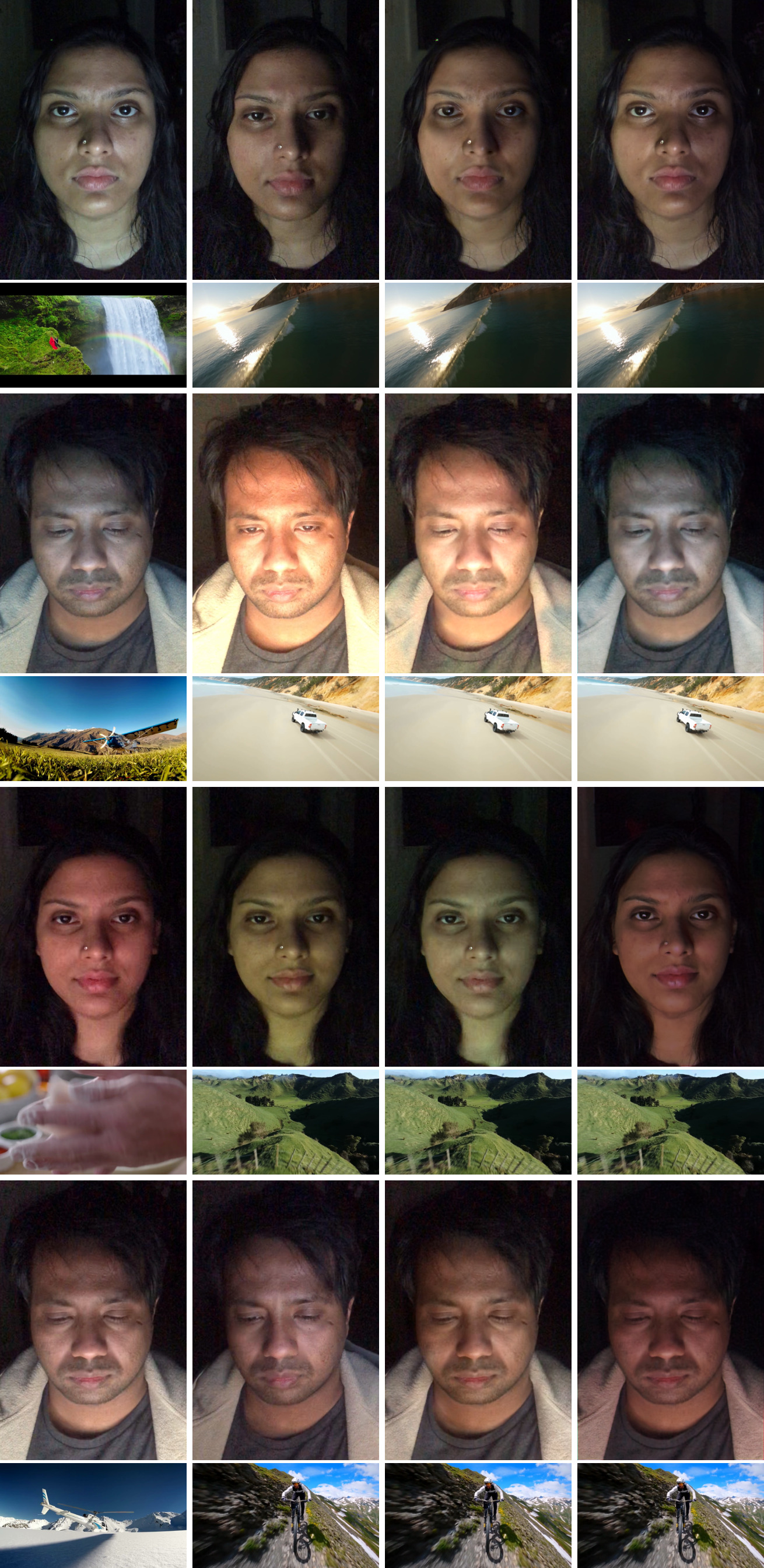}
    \begin{footnotesize}
        \begin{tabularx}{0.6\textwidth}{YYYY}
          Input & Target & Ours & Sun+
        \end{tabularx}
    \end{footnotesize}
    \vspace{-0.5em}
    \caption{\small Qualitative comparison with  Sun+ following Protocol 2 i.e input image and target relighting is not part of the training data.}
    \label{fig:trg2trg5}
    \vspace{-1.0em}
\end{figure*}

\begin{figure*}[h!]
    \centering
    \newcolumntype{A}{>{\hsize=0.202\textwidth}X}
    \newcolumntype{B}{>{\hsize=0.086\textwidth}X}
    \newcolumntype{Y}{>{\centering\arraybackslash}X}
    \includegraphics[width=0.6\textwidth]{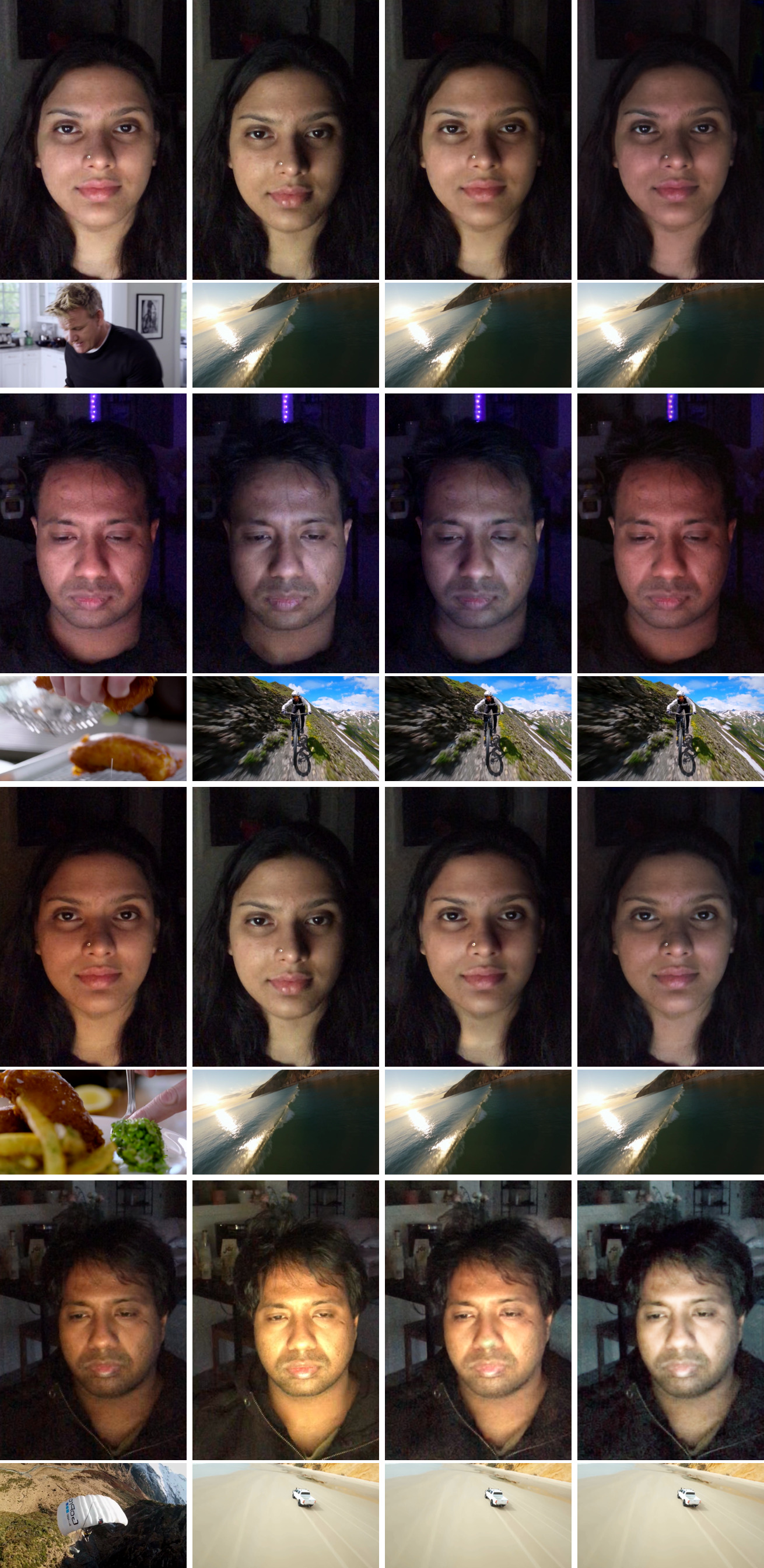}
    \begin{footnotesize}
        \begin{tabularx}{0.6\textwidth}{YYYY}
          Input & Target & Ours & Sun+
        \end{tabularx}
    \end{footnotesize}
    \vspace{-0.5em}
    \caption{\small Qualitative comparison with  Sun+ following Protocol 2 i.e input image and target relighting is not part of the training data.}
    \label{fig:trg2trg6}
    \vspace{-1.0em}
\end{figure*}

\begin{figure*}[h!]
    \centering
    \newcolumntype{A}{>{\hsize=0.202\textwidth}X}
    \newcolumntype{B}{>{\hsize=0.086\textwidth}X}
    \newcolumntype{Y}{>{\centering\arraybackslash}X}
    \includegraphics[width=0.6\textwidth]{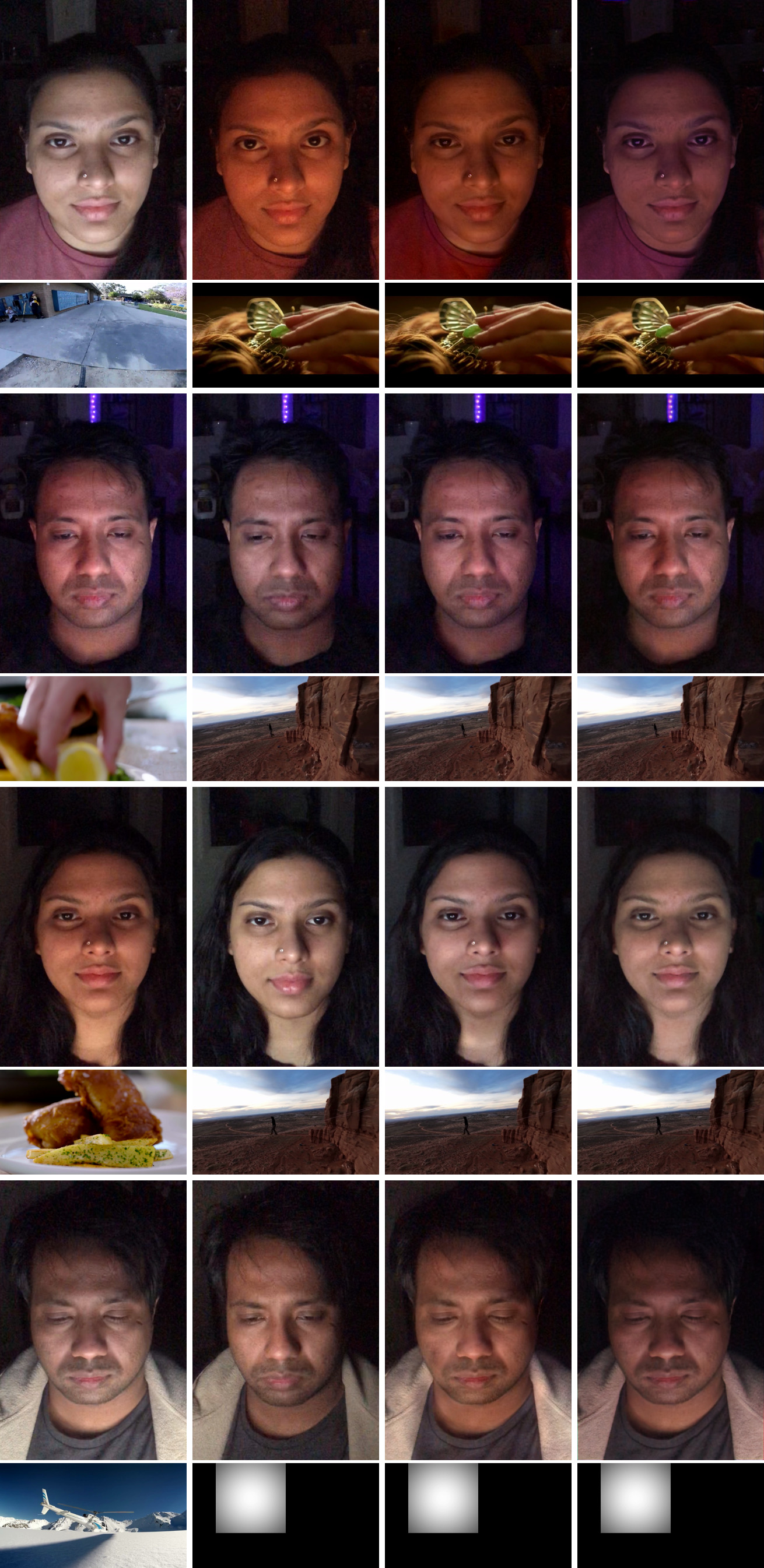}
    \begin{footnotesize}
        \begin{tabularx}{0.6\textwidth}{YYYY}
          Input & Target & Ours & Sun+
        \end{tabularx}
    \end{footnotesize}
    \vspace{-0.5em}
    \caption{\small Qualitative comparison with  Sun+ following Protocol 2 i.e input image and target relighting is not part of the training data.}
    \label{fig:trg2trg7}
    \vspace{-1.0em}
\end{figure*}

We present an overview of additional details and results to be presented in this appendix.
\begin{itemize}
  \setlength\itemsep{0.01cm}
    \item \textbf{Sec \ref{sec:net}} presents details of our network architecture.
    \item In \textbf{Figure \ref{fig:man}} we show visual examples of relighting on a mannequin test sequence. We compare our approach with that of the linear model and RSun. All the methods produce comparable results while the linear method suffers from recovering the correct color. This shows that our implementation of RSun performs as expected in the ideal case.
    \item In \textbf{Figure \ref{fig:lin_fail}} we show that the linear model fails on humans captured with natural head motion.
    \item In \textbf{Figure \ref{fig:src2trg1},\ref{fig:src2trg2},\ref{fig:src2trg3},\ref{fig:src2trg4},\ref{fig:src2trg5},\ref{fig:src2trg6},\ref{fig:src2trg7}} we present additional results comparing our approach to that of Sun+ for Protocol 1 which resembles light stage testing setup, i.e., input image appears in the training data, while target lighting is unseen.
    \item In \textbf{Figure \ref{fig:trg2trg1},\ref{fig:trg2trg2},\ref{fig:trg2trg3},\ref{fig:trg2trg4},\ref{fig:trg2trg5},\ref{fig:trg2trg6},\ref{fig:trg2trg7}} we present  additional results comparing our approach to that of Sun+ for Protocol 2 i.e input image and target relighting is not part of the training data.

\end{itemize}

\subsection{Network Architecture}
\label{sec:net}

Our network architecture, shown in Figure \ref{fig:net}, is based on an U-Net. Our Light encoder consists of a multi-layer perceptron. The light or monitor image is an image of size 18$\times$32, which is converted into a vector. The first layer consists of a fully connected network which predicts a 512 dimensional feature, followed by pixel nomalization and learnable ReLU. The second layer consists of a fully connected network which takes in a 512 dimensional feature and produces a 256 dimensional feature followed by by pixel normalization and learnable ReLU. The same Light Encoder operates on the source and the target monitor image, producing two 256 dimensional features concatenated to produce a 512 dimensional feature. We will release the network architecture code and pre-trained weights.

%%%%%% src2trg Sun+ %%%%%%%

 %%%%%%%%%%%%%%%%%%%%%%%%%%%%%%%%%%%%%%%

%%%%%% trg2trg Sun+ %%%%%%%%%%%%% trg2trg Sun+ %%%%%%%

%%%%%%%%%%%%%%%%

\end{document}